\documentclass[10pt,twocolumn,letterpaper]{article}

\usepackage{cvpr}              %

\usepackage{booktabs}
\usepackage{multirow}
\usepackage{graphicx}  %
\usepackage{amssymb}   %
\usepackage{adjustbox}
\usepackage{xcolor,colortbl}
\usepackage{makecell}

\usepackage{tabularx}
\usepackage{algorithm}
\usepackage{algpseudocode}   %

\makeatletter
\@ifundefined{Cross}{}{}
\makeatother
\usepackage{marvosym}

\usepackage{cuted}
\usepackage{booktabs}  %
\usepackage{array}     %
\usepackage{longtable} %

\renewcommand{\paragraph}[1]{\vspace{.5em}\noindent\textbf{#1}}

\definecolor{cvprblue}{rgb}{0.21,0.49,0.74}
\definecolor{pinkcolor}{RGB}{230, 0, 115}
\usepackage[pagebackref,breaklinks,colorlinks,allcolors=cvprblue]{hyperref}
\hypersetup{
    colorlinks=true,
    urlcolor=pinkcolor,
}

\def\paperID{****} %
\def\confName{CVPR}
\def\confYear{2026}

\title{\raisebox{-0.3\height}{
\includegraphics[height=2.0em]{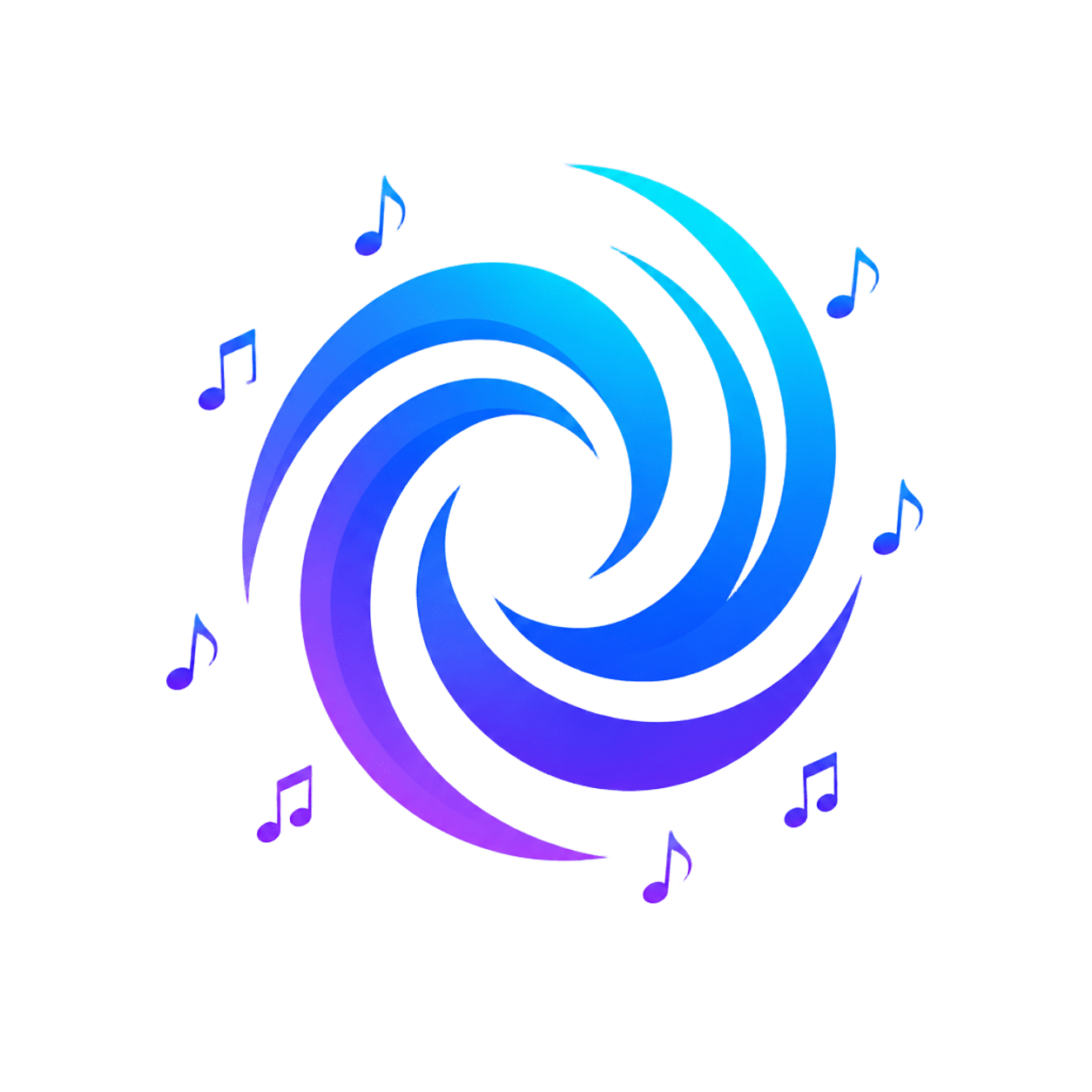}
}
\textbf{UnityShots: Memory-Driven Multi-Shot Audio-Video Generation with Boundary-Aware Gating}
\vspace{-1.2em}
}

\author{
Jiehui Huang\textsuperscript{$1,\dagger$} \quad
Yuechen Zhang\textsuperscript{2} \quad
Bin Xia\textsuperscript{2} \quad
Jiahao Wang\textsuperscript{3} \quad
Xu He\textsuperscript{4} \\
Zhenchao Tang\textsuperscript{5} \quad
Meng Chu\textsuperscript{1} \quad
Xin Tao\textsuperscript{3} \quad
Pengfei Wan\textsuperscript{3} \quad
Jiaya Jia\textsuperscript{1\Letter} \\[0.3em]
\textsuperscript{1}The Hong Kong University of Science and Technology \quad
\textsuperscript{2}The Chinese University of Hong Kong \\
\textsuperscript{3}Kling Team, Kuaishou Technology \quad
\textsuperscript{4}Tsinghua University \quad
\textsuperscript{5}Sun Yat-sen University \\[0.3em]
\url{https://jackailab.github.io/Projects/UnityShots}
}

\newcommand\nnfootnote[1]{%
  \begin{NoHyper}
  \renewcommand\thefootnote{}\footnote{#1}%
  \addtocounter{footnote}{-1}%
  \end{NoHyper}
}

\begin{document}

\twocolumn[{%
\vspace{-0.15in}
\maketitle
\vspace{-0.5in}
\begin{center}
    \captionsetup{type=figure}
    \includegraphics[width=1.0\linewidth]{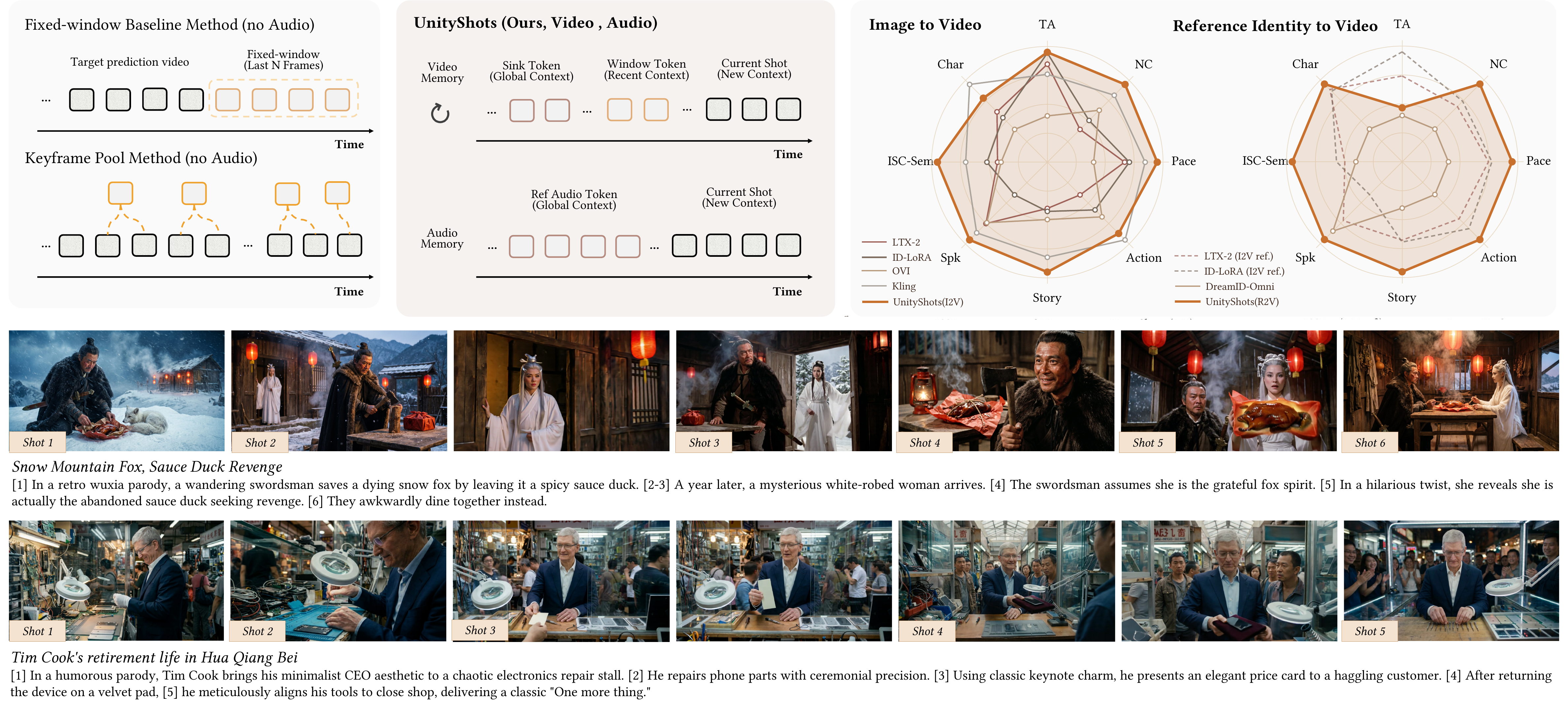}
    \vspace{-0.20in}
    \captionof{figure}{
    \textbf{UnityShots generates consistent multi-shot audio-video sequences across
    diverse cultural contexts.}
    Each row shows shots from the proposed multi-cultural benchmark; the reference
    identity portrait (leftmost) is preserved through hard cuts by the dual-slot memory
    bank and reference audio anchor.
    \emph{Upper left}: UnityShots pairs a long-term (LTM) slot anchored to the opening
    shot with a short-term (STM) slot capturing the preceding tail, plus a cross-shot
    speaker audio anchor, while prior end-to-end and shot-by-shot methods rely on
    single-window context or no persistent cross-shot conditioning.%
    }
    \label{fig_teaser}
\end{center}
\vspace{-0.05in}
}]

\nnfootnote{
\hspace{-2em}$\dagger$ {This work was conducted during the author's internship at Kling Team, Kuaishou Technology.} \\
\hspace{2em}\Letter~{Corresponding Author.}
}

\vspace{-4mm}
\begin{abstract}
\vspace{-8mm}

\noindent Generating a coherent multi-shot video requires structured cross-shot memory. Subject appearance, scene context, and speaker identity must persist across cuts.
Existing approaches either train end-to-end over fixed-length sequences and cannot scale, generate shot-by-shot with memory banks that grow linearly, or orchestrate pretrained generators under an LLM planner without a multi-shot-aware backbone.
We present \textbf{UnityShots}, a memory-driven multi-shot audio-video generation system built on LTX-2.3, trained on annotated cinematic and music-video shots.
The video stream maintains two fixed-size slots, a long-term memory (LTM) slot anchored to the opening shot and a short-term memory (STM) slot holding the immediately preceding tail, both updated at every cut by a boundary-conditioned gate that fuses visual cut probability and beat-tracker signals.
The audio stream injects a reference speaker token at every shot to preserve vocal timbre without a sliding audio bank.
A discrete cut-type prior, learned through AdaLN, becomes an inference-time control knob over transition strength.
We release a benchmark of $200$ multi-cultural multi-shot sequences spanning six ethnic regions and ten or more languages, with per-shot reference identities, reference audio, and per-boundary transition labels.
Evaluated across I2V, T2V, and R2V conditioning modes, UnityShots leads open-source baselines on every cross-shot coherence metric and matches the strongest closed-source system on the multi-shot axes.
Code and data can be found at: \url{https://github.com/JIA-Lab-research/UnityShots}
\end{abstract}

\vspace{-6mm}
\section{Introduction}
\label{sec_intro}
\vspace{-1mm}

Coherent multi-shot video generation requires explicit cross-shot memory, because subject appearance, scene context, and speaker identity must remain stable across shot boundaries.

Existing multi-shot approaches fall into three families, each with a structural
weakness.
The \textbf{end-to-end} family
trains over an entire multi-shot sequence as one denoising
pass~\cite{meng2025holocine, wang2025multishotmaster, wu2025cinetrans, guo2025lct, kara2025shotadapter}, with closed-source systems
such as Kling~\cite{kling2026} and Seedance~\cite{seedance2025} taking the same
direction at much larger scale.
Sequence length is bounded by GPU memory and cost grows with the total clip,
so this family cannot scale to long-form storytelling.
The \textbf{shot-by-shot} family generates each shot conditioned on prior~\cite{luo2026shotstream, zhang2025storymem}, which removes the length cap but introduces
either drift once the conditioning window slides past the opening shot or a keyframe
bank whose memory grows linearly with the sequence.
The \textbf{agent-orchestrated} family pipelines pretrained generators under an LLM
planner~\cite{li2026direct, zhou2026videomemory, liang2025univa}, but the underlying
generator is either not trained for cross-shot coherence (making multi-shot signals hard
to propagate) or accessible only through closed-source APIs.

Despite advances in all three families, current designs treat memory as a
monolithic context window and do not distinguish between what happened long ago and
what happened just before the current cut.
In practice, the next shot's content should integrate both: the opening-shot
identity that anchors the whole narrative and the recent motion and scene state
that determines local continuity.
During long sequences, characters and environments change substantially across shots,
so relying entirely on distant history produces drift, while relying only on the
most recent clip loses the identity anchor.
A second gap is equally important: in real filmmaking and music-video production,
the boundary between shots is driven jointly by the magnitude of visual change and
the rhythm of the audio track.
No existing audio-video generation model treats shot-boundary strength as a joint
function of visual and musical signals, nor adjusts its memory dynamics accordingly.
How to combine audio and visual boundary cues to control long- and short-term memory
in a unified generative model remains an open problem.

We propose \textbf{UnityShots}, a memory-driven multi-shot audio-video generation
system built on LTX-2.3~\cite{hacohen2026ltx}.
Two fixed-size video memory slots---a long-term slot anchored to the opening shot
and a short-term slot holding the immediately preceding tail---are both updated by a
boundary-conditioned gate that fuses visual cut probability and beat-tracker
signals at every shot boundary.
To support this design, we construct a large-scale training set of 146k annotated
cinematic and music-video clips, covering shot-level captions, speaker diarisation,
character anchoring, and transition labels.
The model is trained under three conditioning paradigms---image-to-video (I2V),
text-to-video (T2V), and reference-identity-to-video (R2V)---so it can be steered
by a visual first frame, a text prompt alone, or a reference portrait,
making it suitable for deployment in both manual and agent-driven production pipelines.
A companion agent system, released alongside the model, exposes per-shot generation
as composable tools that an external planner can invoke over sequences of arbitrary
length.

\begin{itemize}[leftmargin=*, itemsep=2pt]
\item \textbf{Memory-driven multi-shot dual-stream generation.}
  An end-to-end-trained I2V/T2V/R2V backbone with a two-slot video memory bank,
  a reference speaker anchor, and a cut-type-aware AdaLN prior that gives users
  explicit control over transition strength at inference time.

\item \textbf{Boundary-conditioned memory gating.}
  A rule-based long-term gate and a content-adaptive short-term gate update the
  memory bank at every cut, jointly driven by visual and audio boundary signals.

\item \textbf{Multi-cultural multi-shot benchmark.}
  200 sequences spanning six cultural groups and ten or more languages, each with
  reference identity images, reference audio, per-shot first frames, narrative
  captions, and per-boundary transition labels.
\end{itemize}

\section{Releated Work}
\label{sec:related}

\subsection{Multi-shot video generation and memory}

Video diffusion models~\cite{ho2022video, ho2022imagen, rombach2022high, blattmann2023stable}
scaled to longer clips; open-source systems~\cite{zheng2024open, yang2025cogvideox} and
large-scale models~\cite{kong2024hunyuanvideo, wan2025wan, brooks2024video, polyak2024movie}
push toward minute-scale generation, but none model shot boundaries as discrete events.

\emph{End-to-end} approaches~\cite{meng2025holocine, wang2025multishotmaster, wu2025cinetrans,
guo2025lct, kara2025shotadapter, xiao2025captaincinema, kling2026, seedance2025} train a single
denoising pass over the full shot sequence, producing coherent clips but bounding sequence
length by GPU memory and training context.
Cinematic design methods~\cite{wang2025cinemaster, xing2025motioncanvas} extend
controllability with 3D-aware scene composition and per-shot camera specification.
\emph{Shot-by-shot} approaches generate each shot conditioned on prior content, lifting
the length cap.
FilmWeaver~\cite{luo2026filmweaver} keeps an autoregressive cache over all prior shots
and STAGE~\cite{zhang2025stage} stores per-entity memory tied to storyboard anchors, but
both grow linearly with the sequence.
OneStory~\cite{an2025onestory} and StoryMem~\cite{zhang2025storymem} select representative
keyframes by appearance similarity, while Cut2next~\cite{he2025cut2next} generates the
next shot via in-context tuning on the preceding clip.
Adapter-based methods~\cite{ye2023ip, zhou2024storydiffusion, huang2024iclora, dahan2026id,
luo2026shotstream} propagate identity features across shots without an explicit temporal
anchor, so identity drift accumulates once the conditioning window slides past the opening
shot.

\begin{figure*}[t]
  \centering
  \includegraphics[width=\linewidth]{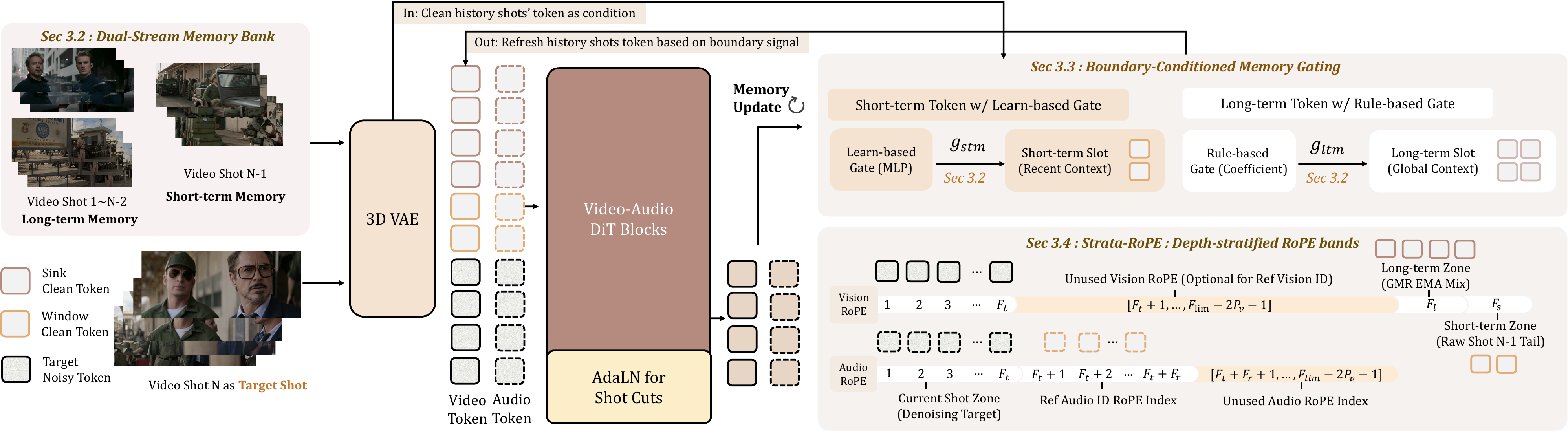}
  \caption{%
    \textbf{UnityShots architecture.}
    \emph{Left}: Strata-RoPE partitions the temporal band of the video stream into three
    non-overlapping strata for the long-term memory (LTM) slot, the short-term memory (STM)
    slot, and the current shot.  
    \emph{Right}: at each shot boundary the composite signal $b_N$ produces two gating
    coefficients $g_{\mathrm{ltm}}$ and $g_{\mathrm{stm}}$ that update the two video
    memory slots before the dual-stream DiT denoises the current shot;
    the audio stream injects a fixed reference speaker token at every shot
    without memory gating.%
  }
  \label{fig:framework}
  \vspace{-6mm}
\end{figure*}

Persistent memory for sequential generation has been studied in language modeling and video
autoregression~\cite{zhang2026frame}, where small fixed-size token sets at reserved positions
act as stable global attractors.
Our bank is recurrent at $O(1)$ cost, updated by a joint audio-visual boundary signal
rather than retrieval, and covers both video and audio streams.
Figure~\ref{fig:framework} (upper left) provides a visual comparison with prior designs.

\subsection{Joint audio-video generation}

Dual-stream architectures~\cite{hacohen2026ltx, guo2026dreamid} show that joint training of
video and audio improves multimodal alignment.
DreamID-Omni~\cite{guo2026dreamid} achieves reference-based audio-video generation with
identity disentanglement but targets single shots and maintains no cross-shot memory.
FoleyCrafter~\cite{zhang2026foleycrafter} demonstrates video-conditioned ambient audio
synthesis from text-to-audio models~\cite{liu2024audioldm}.
None of these approaches maintain a persistent cross-shot speaker anchor; voice
consistency in multi-shot generation relies on per-shot prompting and degrades when
a reference clip is absent.

\subsection{Agent-orchestrated production}

Agent-orchestrated pipelines~\cite{li2026direct, zhou2026videomemory, liang2025univa} wrap
pretrained generators with LLM planners for multi-step video creation, a paradigm gaining
commercial traction in short-drama and music-video production and increasingly viewed as
a practical interface for controllable generative workflows.
The structural limit is that the underlying generator was not trained for cross-shot
coherence, so the planner cannot propagate multi-shot signals cleanly, or it relies on
closed-source paid APIs.
UnityShots is an end-to-end-trained multi-shot backbone compatible with agent-driven
workflows, and a companion agent system exposes per-shot generation as composable tools.

\section{Method}

UnityShots takes a sequence of per-shot prompts together with a reference image and
reference audio, then generates a synchronized multi-shot video-audio clip in which each
shot is denoised conditioned on a structured summary of everything before it.
The backbone is LTX-2.3~\cite{hacohen2026ltx}, a 22B dual-stream diffusion
transformer~\cite{peebles2023scalable} that jointly denoises video and audio latents.
Video tokens use a Block-Relativistic 3D RoPE~\cite{su2024roformer} with temporal
ceiling $f_{\mathrm{lim}} = 128$ indices ($\approx 42$\,s at the native frame rate);
audio tokens use a separate 1-D RoPE with ceiling $f^a_{\mathrm{lim}} = 1024$.
Each shot is generated in image-to-video (I2V) mode, with the first frame as the image
condition, and $P_v = 4$ video frames denote the tail window used to seed memory from a
completed shot.
Given $K$ shots with prompts $\mathbf{c}_0, \ldots, \mathbf{c}_{K-1}$, a reference
identity image $\mathbf{X}^{\mathrm{ref}}$, and a reference audio clip
$\mathbf{A}^{\mathrm{ref}}$, the goal is a multi-shot sequence that maintains consistent
subject appearance and speaker identity across every cut---a problem the backbone has no
native mechanism to solve when shots are generated independently.

\subsection{Dual-Stream Memory Bank}
\label{sec:memdesign}

\paragraph{Video memory bank.}
Visual content changes abruptly at every cut: new scene, new camera angle, new action
phase.
Maintaining identity across cuts therefore requires both a stable long-range anchor and
a fine-grained record of the immediately preceding shot.
The video bank holds two fixed-size slots that serve these two roles.
The \textbf{long-term memory (LTM)} slot $\mathbf{L}^N \in \mathbb{R}^{2 \times C_v}$
uses two latent frames to anchor subject identity to the opening shot.
Its contribution follows a \emph{rule-based} coefficient
(Section~\ref{sec:gating}) that is bounded away from zero at all cut strengths,
so the anchor is never fully erased regardless of what the per-frame detector reports.
The \textbf{short-term memory (STM)} slot $\mathbf{S}^N \in \mathbb{R}^{P_v \times C_v}$
uses $P_v = 4$ latent frames from the immediately preceding tail to capture
action-phase and motion dynamics at the boundary.
Its weight is refined by a small \emph{learned} MLP
(Section~\ref{sec:gating}) conditioned on outgoing and incoming shot content,
so the boundary strength adapts to scene content rather than relying on a fixed schedule.

\paragraph{Audio reference anchor.}
For audio, the cross-shot coherence requirement is more targeted: what must be preserved
is the voice timbre of each character named in the script, not the full spectral trajectory
of the audio track across shots.
We satisfy this by injecting a fixed reference speaker token derived from
$\mathbf{A}^{\mathrm{ref}}$ at every shot; no sliding audio memory window is maintained.
Background music is a structurally separate problem: a musical track evolves continuously
and is not tied to visual edit decisions, so when an application needs accompaniment we
treat it as an external track aligned to the finished video rather than content that must
be co-generated with the visual stream.

\paragraph{Conditioning sequence.}
At shot $N$, the token sequence passed to the backbone is
\begin{equation}
  \mathbf{C}^N = \bigl[\,
    \mathbf{X}^{\mathrm{ref}},\;\;
    \mathbf{L}^N,\;\;
    \mathbf{S}^N,\;\;
    \mathbf{A}^{\mathrm{ref}},\;\;
    \mathbf{V}^N_t \oplus \mathbf{A}^N_t \,
  \bigr],
  \label{eq:context}
\end{equation}
where $\mathbf{V}^N_t \oplus \mathbf{A}^N_t$ is the noisy video-audio latent at diffusion
timestep $t$.
The denoising loss is masked to the final block only, so the model never predicts memory
tokens.
All five blocks are concatenated along the token dimension and processed by the shared
dual-stream attention~\cite{vaswani2017attention} in a single forward pass.

\subsection{Boundary-Conditioned Memory Gating}
\label{sec:gating}

\paragraph{Cut-type prior and boundary score.}
Each training shot is labeled with a discrete transition type
$\tau_N \in \{\textsc{first}, \textsc{continue}, \textsc{hard}\}$ with priors
$\{0, 0.4, 1.0\}$.
A small embedding of $\tau_N$ is fused with the diffusion timestep and injected into
every DiT block via AdaLN; at inference, setting $\tau_N$ directly controls
how strongly the next cut resets the memory.
Within each transition class, we compute a continuous boundary score
\begin{equation}
  b_N = \tau_N \cdot \bigl(\alpha \, s_{\mathrm{vis}}
                          + \beta \, s_{\mathrm{aud}}
                          + \gamma \, s_{\mathrm{beat}}\bigr),
  \label{eq:boundary}
\end{equation}
where $s_{\mathrm{vis}}$ is the TransNetv2~\cite{soucek2020transnet} visual-cut probability,
$s_{\mathrm{aud}}$ an RMS energy-change score, and
$s_{\mathrm{beat}} \in \{0.25, 0.5, 1.0\}$ encodes beat position from a beat
tracker~\cite{foscarin2024beat}.
The prior acts as an upper envelope: a \textsc{continue} shot never over-resets the
LTM anchor even if the visual detector misfires.

\paragraph{Memory update.}
From the clipped boundary score $\bar{b}_N = \mathrm{clamp}(b_N, 0, 1)$, we
derive two monotone-increasing gating coefficients $g_{\mathrm{ltm}}(\bar{b}_N)$
and $g_{\mathrm{stm}}(\bar{b}_N)$ on non-overlapping ranges (values given in
Section~\ref{sec:impl}).
The LTM coefficient is bounded strictly above zero, while the STM coefficient
reaches near-full weight at a hard cut to bridge the boundary.
Since the LTM slot contains only $P_{\ell} = 2$ latent frames, we use
$h_{\ell}(\cdot)$ to select the last $P_{\ell}$ latent frames from the
previous-shot tail before writing to LTM.
Both slots are updated before each denoising pass:
\begin{equation}
  \mathbf{L}^N \leftarrow
    z_N\,h_{\ell}(\mathbf{V}^{N-1}_{\mathrm{tail}})
    + (1 - z_N)\,\mathbf{L}^{N-1},
  \qquad
  \mathbf{S}^N \leftarrow \mathbf{V}^{N-1}_{\mathrm{tail}},
  \label{eq:gmr}
\end{equation}
with $z_N = z_{\max}\bar{b}_N$ and $z_{\max} < 1$, then scaled by
$g_{\mathrm{ltm}}(\bar{b}_N)$ and $g_{\mathrm{stm}}(\bar{b}_N)$ before entering
Equation~\ref{eq:context}.
This bounded update prevents the opening-shot anchor from being abruptly
overwritten.

\paragraph{Content-aware refinement.}
A two-layer MLP ($<50$K parameters) adds a content-aware correction to $g_{\mathrm{stm}}$:
it takes pooled features from the outgoing and incoming shots together with the transition
embedding and produces a multiplicative factor, clamped to $[0.5, 1.0]$ for the first
$500$ training steps to prevent collapse.
The LTM gate stays rule-based as shot~$0$ has no prior content to condition on.

\subsection{Strata-RoPE Position Encoding}
\label{sec:strataRoPE}

Attention must distinguish LTM content, STM content, and the current noisy shot from
each other without extra depth tokens or side networks.
We assign each video block a fixed stratum along the temporal RoPE axis:
\begin{equation}
  \begin{aligned}
  \text{current shot} &\;:\; [0,\; T_v - 1], \\
  \text{STM band} &\;:\; [f_{\mathrm{lim}} - P_v,\; f_{\mathrm{lim}} - 1], \\
  \text{LTM band} &\;:\; [f_{\mathrm{lim}} - 2P_v,\; f_{\mathrm{lim}} - P_v - 1].
  \end{aligned}
  \label{eq:strataRoPE}
\end{equation}
The current shot and the memory strata are far apart on the integer axis, so the rotary
kernel attenuates cross-strata interaction in proportion to distance automatically.
The audio stream uses standard reference-token injection and does not need a multi-tier
RoPE layout, since a single anchor at every shot has no prior audio content to separate
from positionally.
This differs from the negative-offset keyframe layout of StoryMem~\cite{zhang2025storymem},
which encodes the absolute time of a memory frame. Our encoding marks \emph{which stratum}
a token occupies, a design suited to the recurrent structure of the memory bank.

\subsection{Training and Inference}
\label{sec:trainingstrategy}

\paragraph{Training.}
Stage~1 fine-tunes the backbone on single shots with the reference block only.
Stage~2 adds both memory slots and boundary-conditioned gating, trains on $k$-shot
chunks ($k\!\in\![3,9]$) in mixed conditioning mode: image-to-video ($p\!=\!0.5$),
text-to-video ($p\!=\!0.3$), and reference-identity-to-video ($p\!=\!0.2$).
The recurrence is unrolled over the chunk; the diffusion loss~\cite{ho2020denoising}
\begin{equation}
  \mathcal{L} = \mathbb{E}_{\mathbf{V}^N,\,t,\,\boldsymbol{\epsilon}}\bigl[
    \|\boldsymbol{\epsilon} - \boldsymbol{\epsilon}_\theta(\mathbf{C}^N, t)\|_2^2
  \bigr]
  \label{eq:loss}
\end{equation}
is masked to the final shot only.
The audio branch is fine-tuned jointly but does not use the cross-shot memory bank.
The only cross-shot audio signal is the reference speaker token (Section~\ref{sec:memdesign}).
Random reference-audio dropout ($p\!=\!0.1$) allows text-only fallback.

\paragraph{Inference.}
For each shot in sequence, the STM slot receives the $P_v$-frame tail of the previous
shot, the LTM slot is updated by Eq.~\ref{eq:gmr}, both are modulated by their gating
coefficients, and the backbone denoises the current shot conditioned on
Eq.~\ref{eq:context}.
The R2V variant, which replaces the per-shot first frame with a concatenated reference identity token, relaxes the first-frame conditioning constraint and enables more flexible generation.

\section{Experiment}

\subsection{Implementation Details}
\label{sec:impl}

Both training stages use AdamW with learning rate $1\times10^{-5}$ (gate MLP: $5\times10^{-5}$),
gradient clipping at $1.0$, BFloat16 precision, and DeepSpeed~\cite{rajbhandari2020zero} on $4$ nodes of $8$
NVIDIA A800 GPUs each.
Stage~1 fine-tunes for 20{,}000 steps on single-shot identity data.
Stage~2 trains on multi-shot chunks with variable length $k\!\in\![3, 9]$ sampled by
a bucketed-length sampler, in mixed conditioning mode (image-to-video $p\!=\!0.5$,
text-to-video $p\!=\!0.3$, reference-identity-to-video $p\!=\!0.2$); $30\%$ of
samples use a subtitle-free variant to match the base model's distribution.
Evaluation uses DDIM~\cite{song2020denoising}-based inference with $50$ steps.

The boundary-conditioned coefficients of Section~\ref{sec:gating} are set to
$g_{\mathrm{ltm}}(b) = 0.1 + 0.6\,b$ and $g_{\mathrm{stm}}(b) = 0.3 + 0.7\,b$,
selected by a small grid search over
$\{0.0, 0.1, 0.2\} \times \{0.4, 0.6, 0.8\}$ on a held-out validation split.
Detailed hyperparameters are listed in Appendix~A.

We compare four configurations to isolate each design axis:
\textbf{LTX-2.3 (no memory)} runs the unmodified backbone independently per shot;
\textbf{+\,STM only} adds just the short-term memory slot with the boundary-conditioned
gating but no long-term anchor;
\textbf{+\,LTM only} adds just the long-term memory slot with the boundary-conditioned
gating but no short-term window;
\textbf{UnityShots (full)} uses both memory slots together with the full
boundary-conditioned gating from Section~\ref{sec:gating} and the reference speaker
anchor for audio.

\subsection{Datasets, Benchmark, and Metrics}
\label{sec:datasets}

\paragraph{Training data.}
Stage~2 uses 116k cinematic clips (segmented by TransNetv2~\cite{soucek2020transnet}) and 30k
music-video clips (segmented by a beat tracker~\cite{foscarin2024beat} with beat-strength labels),
totaling 146k shots.

\begin{figure*}[p]
  \centering
  \includegraphics[width=\linewidth]{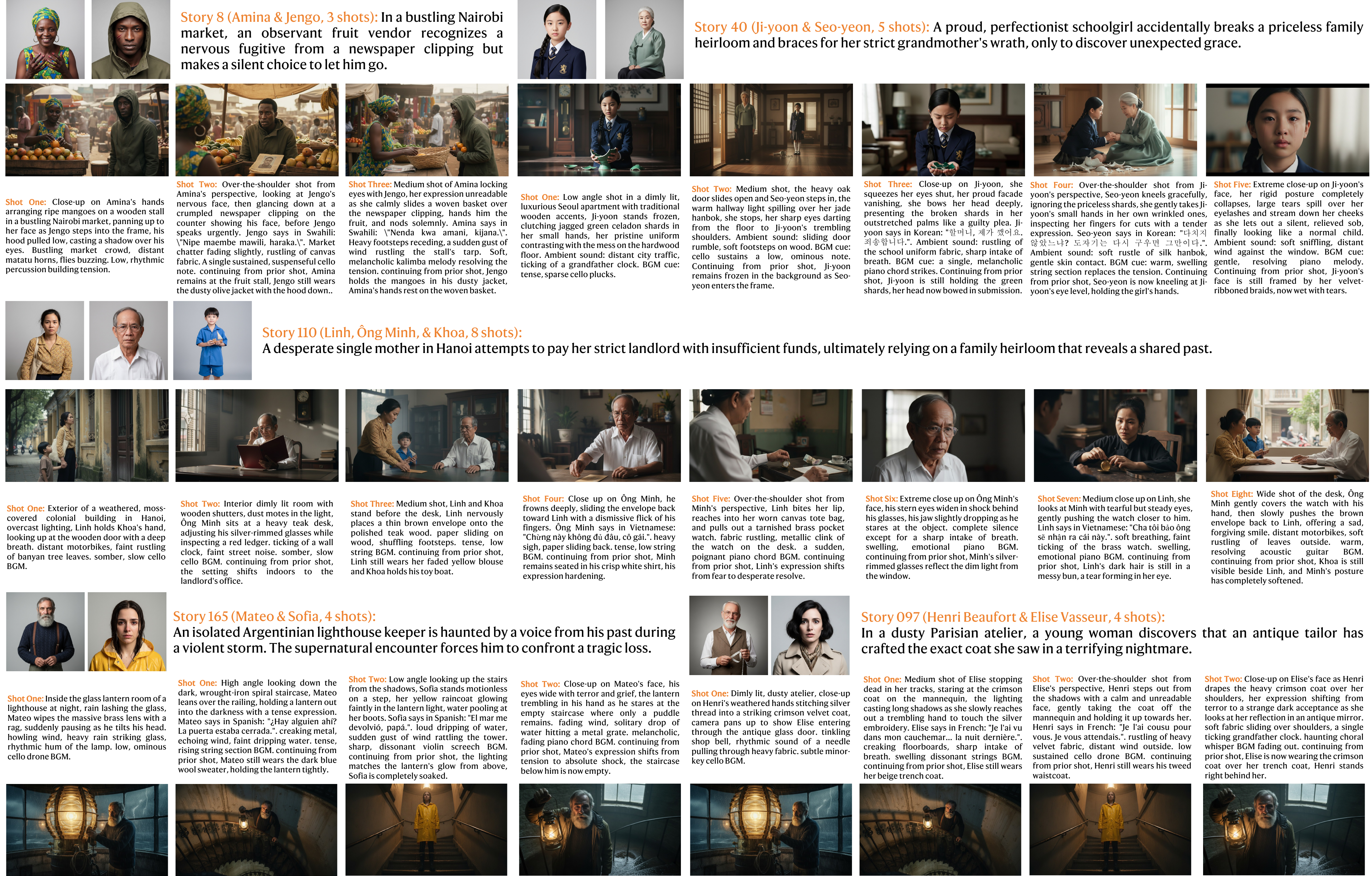}
  \caption{%
    \textbf{Multi-cultural benchmark overview.}
    Representative reference portraits, per-shot first frames, and generated sequences
    from six ethnic regions covered by the $200$-sequence benchmark.
    Each row shows a distinct cultural context; characters share identity and voice
    across all shots in a sequence.
  }
  \label{fig:benchmark_overview}
\end{figure*}

\paragraph{Evaluation benchmark.}
We release $200$ multi-shot sequences from a multi-cultural short-drama pipeline covering
six cultural groups and ten or more languages (Figure~\ref{fig:benchmark_overview}).
Each sequence has $3$--$9$ shots with unique story outlines, reference identity images,
reference audio clips, per-shot first frames, and per-boundary transition labels.
Full construction details are in Appendix~C (Figure~\ref{fig:data_pipe}).

\paragraph{Metrics.}
\textbf{TA}: text-video alignment scored by VBench~\cite{huang2024vbench} using ViCLIP~\cite{wang2024internvid} ($\times 100$).
\textbf{TSIM}: DINOv2~\cite{oquab2023dinov2} cosine similarity between consecutive-shot
frames, following the inter-shot consistency protocol of StoryMem~\cite{zhang2025storymem}.
\textbf{AES-V}/\textbf{AES-A}: VBench visual aesthetics and Audiobox~\cite{tjandra2025meta}
audio aesthetics.
\textbf{CLAP}~\cite{elizalde2023clap}: audio--caption similarity averaged across shots.
\textbf{NC}, \textbf{Story}, \textbf{Char}, \textbf{Pace}, \textbf{Cult}: Gemini-2.5-Pro
$1$--$5$ ratings ($8$ frames/sequence, \texttt{temperature=0}); Cult uses a separate prompt
conditioned on per-shot dialogue and character ethnic context.

\begin{table*}[t]
\centering
\caption{%
  Multi-shot generation on the multi-cultural benchmark grouped into
  video, audio, and multi-shot coherence axes (I2V / T2V / R2V conditioning).
  $\uparrow$ higher is better. \textbf{Bold} and \underline{underline} mark
  first and second place within each conditioning block.
  HoloCine produces no audio (shown as $-$).%
}
\label{tab:competitor}
\setlength{\tabcolsep}{4.6pt}
\renewcommand{\arraystretch}{1.18}
\footnotesize{
\begin{tabular*}{\textwidth}{@{\extracolsep{\fill}}llccc|cc|cccc@{}}
\toprule
\multirow{2}{*}{Method} & \multirow{2}{*}{Input}
  & \multicolumn{3}{c|}{\textit{Video}}
  & \multicolumn{2}{c|}{\textit{Audio}}
  & \multicolumn{4}{c}{\textit{Multi-shot Coherence}} \\
\cmidrule(lr){3-5}\cmidrule(lr){6-7}\cmidrule(lr){8-11}
 & & TA $\uparrow$ & TSIM $\uparrow$ & AES-V $\uparrow$
  & AES-A $\uparrow$ & CLAP $\uparrow$
  & NC $\uparrow$ & Story $\uparrow$ & Char $\uparrow$ & Pace $\uparrow$ \\
\midrule
\multicolumn{11}{l}{\textit{I2V — image-to-video conditioning}} \\
LTX-2~\cite{hacohen2026ltx}              & I2V & 16.75 & \underline{0.370} & 0.508 & 7.11 & 0.095 & 3.53 & 3.28 & 3.91 & 3.72 \\
ID-LoRA~\cite{dahan2026id}                & I2V & 16.39 & 0.362 & 0.507 & 7.01 & 0.113 & 3.80 & 3.45 & 4.04 & 3.45 \\
OVI~\cite{low2025ovi}                       & I2V & 15.86 & 0.323 & 0.568 & 6.37 & 0.162 & 3.44 & 2.97 & 4.02 & 3.79 \\
MOVA~\cite{team2026mova}                 & I2V & 16.13 & 0.298 & \underline{0.575} & 6.54 & 0.146 & 3.98 & 3.56 & 4.45 & \underline{4.18} \\
Kling~\cite{kling2026}                   & I2V & 19.20 & 0.378 & \textbf{0.610} & \underline{6.83} & \underline{0.155} & \underline{4.25} & \underline{4.15} & \textbf{4.75} & \textbf{4.20} \\
\textbf{UnityShots~(I2V) (ours)}          & I2V & \textbf{20.62} & \textbf{0.392} & 0.563 & \textbf{7.30} & \textbf{0.170} & \textbf{4.38} & \textbf{4.25} & \underline{4.54} & \textbf{4.20} \\
\midrule
\multicolumn{11}{l}{\textit{T2V — text-to-video conditioning}} \\
HoloCine~\cite{meng2025holocine}             & T2V & 18.16 & 0.401    & 0.494 & --   & --    & 3.51 & 3.22 & 3.72 & \underline{3.28} \\
\textbf{UnityShots~(T2V) (ours)}          & T2V & \textbf{19.17} & \textbf{0.451} & \textbf{0.540} & \textbf{7.39} & \textbf{0.186} & \textbf{4.13} & \textbf{3.83} & \textbf{4.11} & \textbf{3.39} \\
\midrule
\multicolumn{11}{l}{\textit{R2V — reference-identity-to-video conditioning}} \\
DreamID-Omni~\cite{guo2026dreamid}      & R2V & 16.81 & 0.490 & \textbf{0.555} & 6.56 & 0.168 & 2.84 & 2.64 & 3.14 & 2.90 \\
\textbf{UnityShots~(R2V) (ours)}          & R2V & \textbf{17.98} & \textbf{0.548} & 0.543 & \textbf{7.57} & \textbf{0.176} & \textbf{3.36} & \textbf{3.12} & \textbf{3.40} & \textbf{3.44} \\
\bottomrule
\end{tabular*}
}
\end{table*}

\subsection{Qualitative Results}
\label{sec:qualitative}

\begin{figure*}[t]
  \centering
  \includegraphics[width=\linewidth]{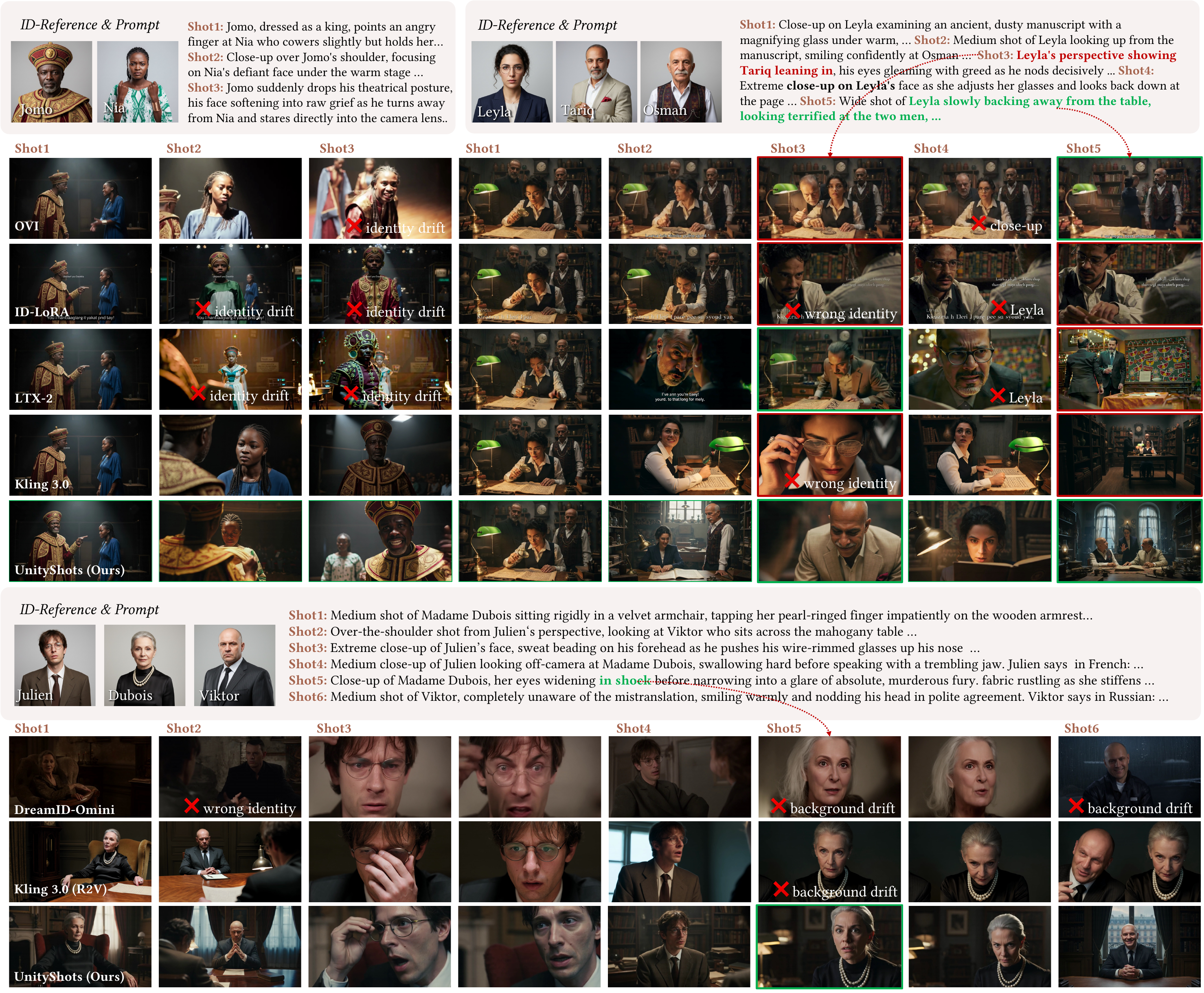}%
  \caption{%
    \textbf{Qualitative comparison on the multi-cultural benchmark.}
    Columns show LTX-2 (no memory), ID-LoRA, Kling, and UnityShots(I2V) on 3-shot and
    5-shot sequences from diverse cultural contexts.
    UnityShots follows per-shot prompts while maintaining stable subject identity across
    all cuts. LTX-2 and ID-LoRA accumulate visible drift from shot~$2$ onward, and Kling,
    despite high per-frame fidelity, shows subject-framing inconsistencies across cuts
    that the LTM anchor prevents by maintaining a persistent latent encoding of the
    opening-shot reference through every denoising pass.
  }
  \vspace{-3mm}
  \label{fig:compared}
\end{figure*}

\paragraph{Short sequences (3 and 5 shots) against open-source baselines.}
As shown in Fig.~\ref{fig:compared}, on 3-shot and 5-shot I2V sequences, UnityShots(I2V)
generates each shot with higher fidelity to the per-shot text prompt than open-source
baselines while maintaining stable subject identity across cuts.
LTX-2.3 without memory and ID-LoRA both accumulate visible drift from shot~$2$
onward: identity attributes such as hairline, clothing, and skin tone shift relative
to the reference portrait even when the prompt is unchanged.

\paragraph{Comparison against the closed-source Kling.}
Kling demonstrates superior single-shot visual quality from large-scale training, but
exhibits cross-shot consistency gaps on sequences requiring consistent subject framing
under different camera angles.
UnityShots leads on narrative coherence (NC) and cinematic pacing (Pace) because the
LTM slot anchors the opening shot throughout the sequence.

\paragraph{Longer sequences (6 shots and beyond).}
On longer sequences, identity drift is the dominant failure mode: clothing, face details,
and scene framing diverge from the opening reference once no persistent cross-shot
anchor exists.
UnityShots avoids this collapse through the dual-tier memory: the LTM slot supplies
a persistent opening-shot encoding to every denoising pass while the STM slot bridges
immediate-boundary transitions in motion and lighting.
The rule-based gate keeps the LTM contribution non-zero even at shot~8; quantitative
gains across length brackets are in Table~\ref{tab:longseq}.

\subsection{Main Results}
\label{sec:main}

Table~\ref{tab:competitor} compares UnityShots against baselines on
the multi-cultural benchmark across three conditioning modes.
The I2V section evaluates open-source, closed-source
Kling~\cite{kling2026}, and UnityShots(I2V).
The T2V section compares HoloCine~\cite{meng2025holocine} against UnityShots(T2V).
The R2V section compares DreamID-Omni~\cite{guo2026dreamid} against UnityShots(R2V).
All methods are evaluated on same benchmark sequences.

\paragraph{I2V comparison.}
UnityShots(I2V) leads every video and multi-shot coherence metric against open-source
I2V baselines; the gap is sharpest on cross-shot axes: $+0.40$ NC and $+0.69$ Story
over the strongest open-source baseline, with AudioAES-A and CLAP also best due to the
reference-audio cross-shot anchor.
Against Kling, which leads on per-frame fidelity (AES-V $0.610$, Char $4.75$) from its
larger training scale, UnityShots still leads on NC ($+0.13$), Story ($+0.10$), and
every audio metric (AES-A $+0.47$), since Kling produces ambient audio rather than
speaker-anchored cross-shot audio.

\begin{figure*}[p]
  \centering
  \includegraphics[width=\linewidth]{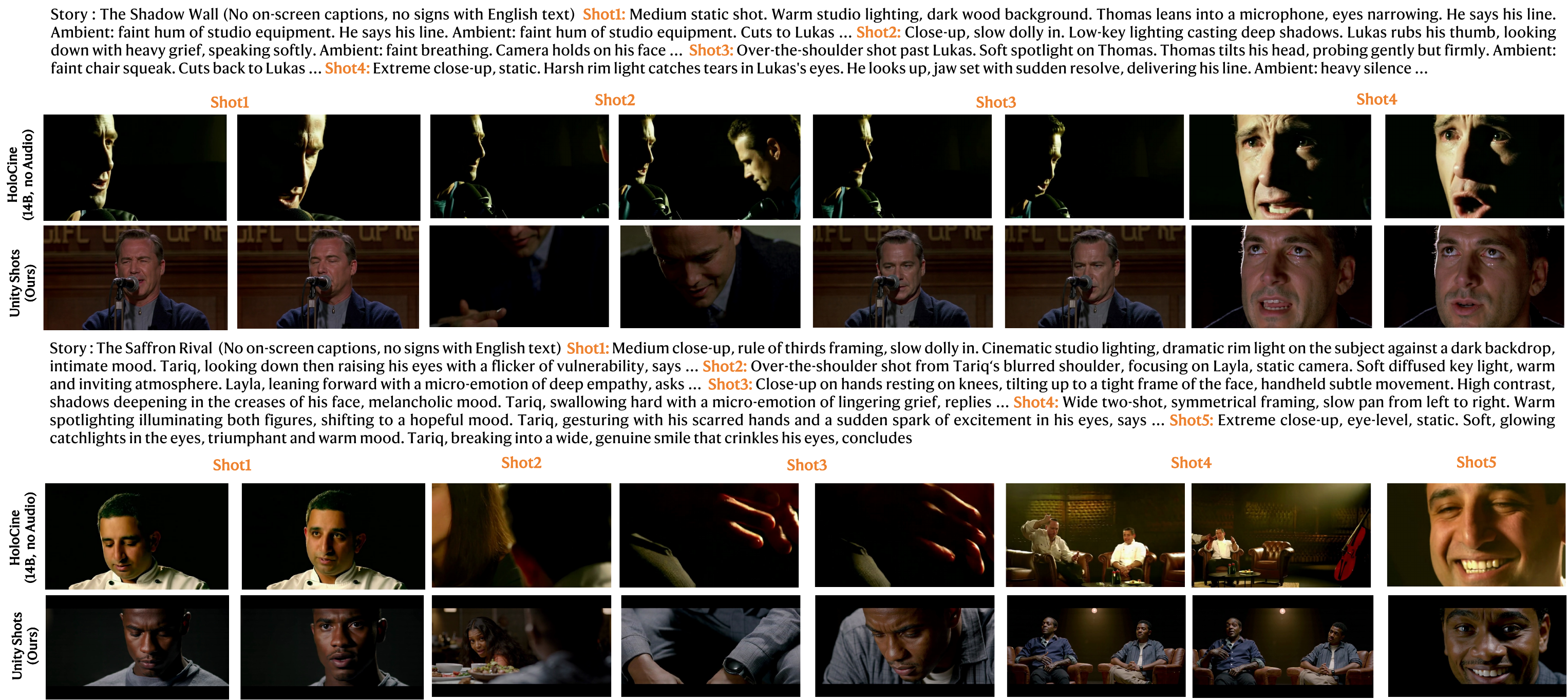}
  \caption{%
    \textbf{Qualitative comparison in T2V mode.}
    Each column shows a different method generating a 4-shot sequence from text prompts alone.
    UnityShots(T2V) maintains consistent subject appearance and scene context across all shots
    while producing synchronized audio.
    HoloCine achieves competitive per-shot visual quality but lacks a cross-shot memory
    mechanism and generates no audio track.
    Open-source T2V baselines exhibit visible identity and background drift from shot 2 onward.
  }
  \label{fig:qualitative}
\end{figure*}

\paragraph{T2V and R2V comparison.}
As shown in Figure~\ref{fig:qualitative}, under the T2V protocol UnityShots(T2V)
outperforms HoloCine~\cite{meng2025holocine} on every shared metric ($+0.62$ NC,
$+0.61$ Story) and additionally provides synchronised audio that HoloCine does not generate.
In R2V mode, UnityShots(R2V) outperforms DreamID-Omni~\cite{guo2026dreamid} on
$8$ of $9$ metrics on a matched $50$-sequence subset: the largest gaps are on the
multi-shot-coherence axes ($+0.52$ NC, $+0.48$ Story, $+0.54$ Pace) and on the
audio block (AES-A $+1.01$, CLAP $+0.008$), reflecting DreamID-Omni's lack of any
cross-shot temporal or audio anchor.
The only column where DreamID-Omni leads is AES-V ($0.555$ vs.\ $0.543$),
consistent with its identity-focused per-frame fidelity at the cost of cross-shot
narrative.

\paragraph{Fairness via the I2V-QwenIMG protocol.}
A confound in chained-I2V evaluation is that baselines which forward the previous
shot's last frame as the next first frame accumulate identity drift independent of
the multi-shot model.
To remove this confound, Table~\ref{tab:qwen_protocol} re-runs the comparison
with all per-shot first frames generated by a single Qwen-Image-Edit-2511~\cite{wu2025qwenimagetechnicalreport} model
conditioned on the reference identity portrait and the per-shot caption so every
method sees an identical first-frame distribution.
UnityShots(I2V) retains its lead on narrative metrics under this setting;
additionally, using the NanoBanana image generator instead of QwenImage produces higher
overall visual quality for all methods while preserving the ranking.
Additional comparisons against open-source video benchmarks are provided in Appendix~E.

\begin{table}[h]
\centering
\caption{%
  Results under two controlled first-frame sources on a held-out subset.
  NanoBanana uses a high-quality image generator; QwenImage uses Qwen-Image-Edit-2511
  conditioned on the reference portrait and per-shot caption.
  $\uparrow$ higher is better. \textbf{Bold} and \underline{underline} mark
  first and second within each frame-source block.%
}
\label{tab:qwen_protocol}
\footnotesize
\setlength{\tabcolsep}{6pt}
\begin{tabular}{lc@{\hspace{6pt}}ccccc}
\toprule
Method & Frame & NC $\uparrow$ & Pace $\uparrow$ & Story $\uparrow$ & Char $\uparrow$ & Cult $\uparrow$ \\
\midrule
LTX-2         & \multirow{5}{*}{\makecell[c]{Nano\\Banana}} & 4.19 & 3.91 & 3.82 & 4.45 & 3.03 \\
ID-LoRA       &                            & 4.47 & 3.73 & 4.36 & 4.73 & 3.06 \\
OVI           &                            & 4.55 & \underline{3.91} & 4.36 & \underline{5.00} & 3.58 \\
MOVA          &                            & \underline{4.78} & 3.82 & \underline{4.55} & 4.91 & \underline{3.64} \\
\textbf{UnityShots~(I2V)} &                 & \textbf{4.82} & \textbf{4.55} & \textbf{5.00} & \textbf{5.00} & \textbf{3.70} \\
\midrule
LTX-2         & \multirow{5}{*}{\makecell[c]{Qwen\\Image}}   & 4.17 & 2.36 & 3.73 & 4.36 & 2.15 \\
ID-LoRA       &                            & 4.45 & 1.64 & 4.27 & 4.64 & 1.70 \\
OVI           &                            & 4.52 & \textbf{2.73} & 4.27 & \underline{4.89} & \underline{2.79} \\
MOVA          &                            & \underline{4.75} & \underline{2.36} & \underline{4.45} & 4.81 & \textbf{2.94} \\
\textbf{UnityShots~(I2V)} &                 & \textbf{4.82} & 2.18 & \textbf{4.93} & \textbf{4.96} & 2.30 \\
\bottomrule
\end{tabular}
\end{table}

\subsection{Ablation Study}
\label{sec:ablation}

\begin{table}[h]
\centering
\caption{%
  Component ablation in I2V (top) and R2V (bottom) conditioning modes
  on a held-out subset of the benchmark.
  $\uparrow$ higher is better. \textbf{Bold} and \underline{underline} mark
  first and second per column within each block.%
}
\label{tab:ablation}
\footnotesize
\setlength{\tabcolsep}{9pt}
\begin{tabular}{lcccc}
\toprule
Configuration & NC $\uparrow$ & Story $\uparrow$ & Char $\uparrow$ & Cult $\uparrow$ \\
\midrule
\multicolumn{5}{l}{\textit{I2V — image-to-video}} \\
LTX-2.3 (no memory)             & \underline{4.45} & 3.60 & \underline{4.40} & \textbf{4.20} \\
+\,STM only                     & 4.30 & \underline{4.04} & 4.18 & 2.93 \\
+\,LTM only                     & 3.95 & 3.97 & 4.22 & \underline{3.27} \\
\textbf{UnityShots~(I2V) (ours)} & \textbf{4.95} & \textbf{4.80} & \textbf{5.00} & \underline{3.33} \\
\midrule
\multicolumn{5}{l}{\textit{R2V — reference-identity-to-video}} \\
+\,STM only                     & \underline{3.01} & \underline{2.75} & \underline{3.20} & 2.63 \\
+\,LTM only                     & 2.84 & 2.53 & 3.13 & \textbf{2.77} \\
\textbf{UnityShots~(R2V) (ours)} & \textbf{3.16} & \textbf{2.90} & \textbf{3.32} & \underline{2.69} \\
\bottomrule
\end{tabular}
\end{table}

\begin{figure}[t]
  \centering
  \includegraphics[width=\linewidth]{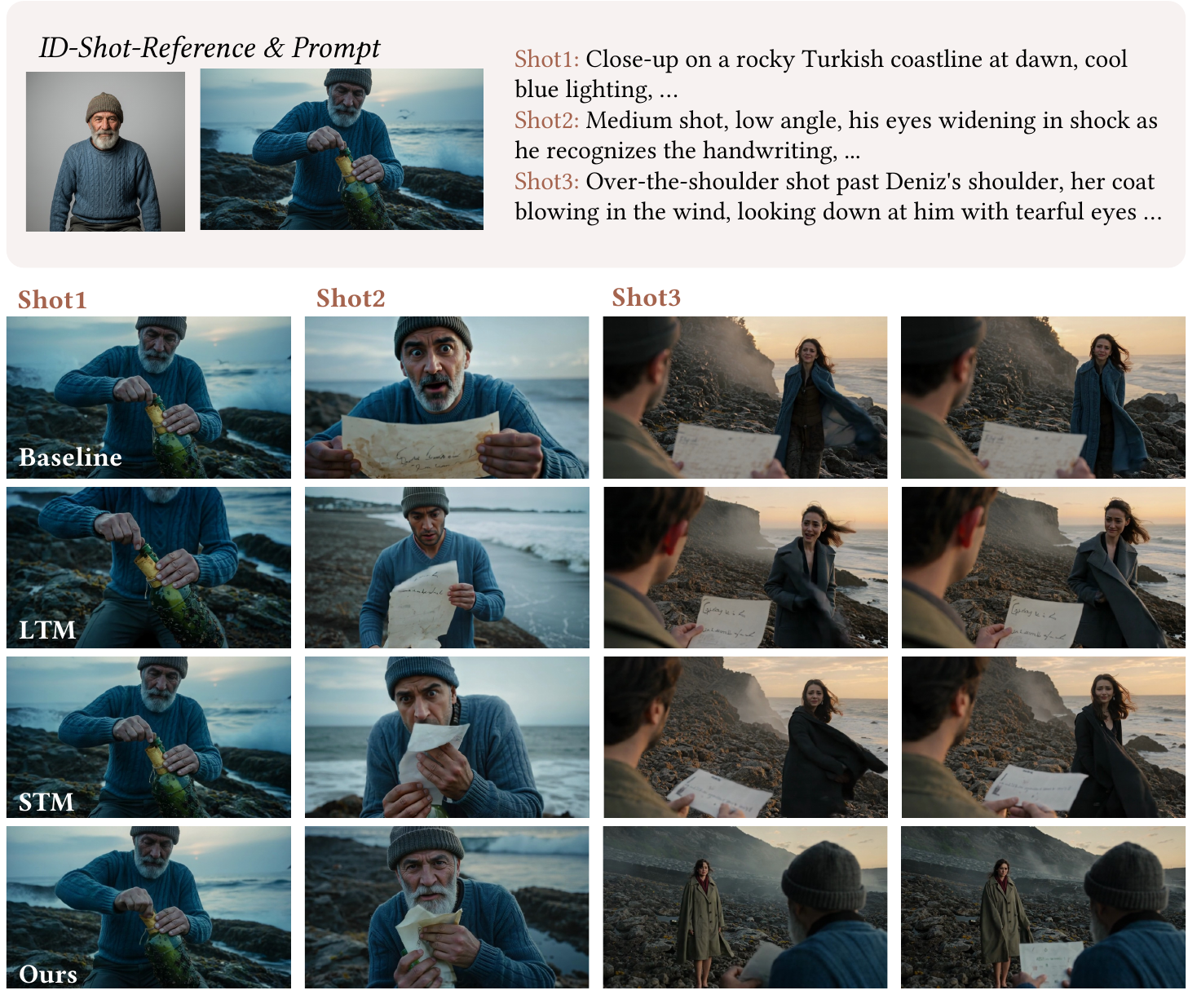}
  \caption{%
    Ablation visualization on representative multi-cultural sequences.
    Columns show the no-memory baseline, STM-only, and the full system;
    the full system maintains subject appearance across all four shots while
    each single-slot variant drifts.%
  }
  \vspace{-6mm}
  \label{fig:aba_vis}
\end{figure}

\paragraph{Both memory slots are necessary.}
Table~\ref{tab:ablation} shows that activating either slot \emph{alone} lowers NC
relative to the no-memory backbone (STM-only $4.30$, LTM-only $3.95$,
vs.\ no-memory $4.45$) while improving Story and Char, revealing a complementarity
requirement.
STM alone provides local-continuity cues without a long-range identity anchor;
LTM alone supplies that anchor but lacks recent visual context for smooth transitions.
Only when both are active do all four metrics simultaneously exceed no-memory,
confirming their complementary roles
(Figure~\ref{fig:aba_vis}).
In R2V mode the full system gains $+0.15$ NC and $+0.12$ Char over the best
single-slot variant, where LTM is the only persistent identity reference available.

\paragraph{Cultural authenticity and memory.}
The no-memory baseline leads on Cult in the I2V ablation ($4.20$ vs.\ $3.27$--$3.33$
for memory-augmented variants), revealing that fixed identity tokens can suppress
cultural-specific visual details the backbone would otherwise infer from the prompt.
This trade-off is discussed further in Section~\ref{sec:conclusion}.

\paragraph{Effect of LTM gating strategy.}
Replacing the rule-based LTM coefficient with a learned MLP gate (same architecture as the
STM gate) performs comparably on 3--4 shot sequences but degrades at $6+$ shots
($-0.29$ NC, $-0.50$ Story) as gradient updates gradually suppress the opening-shot
contribution.
A fixed monotone-in-$b_N$ schedule is immune to this suppression; full results by
sequence length are in Appendix~H.

\subsection{Long-Sequence Robustness}
\label{sec:longseq}

The fixed monotone LTM gate keeps a non-zero opening-shot contribution regardless of
sequence length.
Figure~\ref{fig:longseq}(a) shows UnityShots leading at early shot indices and ahead
at most later ones. Figure~\ref{fig:longseq}(b) confirms the cumulative NC advantage
grows with the shot horizon, ending $+0.11$ above the strongest baseline.
Table~\ref{tab:longseq} breaks this down across three length brackets.
At $3$--$4$ shots all methods are close; at $5$ shots the STM slot visibly suppresses
the identity drift that baselines accumulate; at $6+$ shots MOVA represents the
strongest baseline ($4.71$ NC, $4.50$ Story) while UnityShots reaches a perfect
NC/Story/Char rating from the Gemini evaluator, the subset where memory has the
largest perceptual impact.

\begin{table}[!t]
\centering
\caption{%
  Long-sequence robustness across shot-count brackets.
  $\uparrow$ higher is better. \textbf{Bold} and \underline{underline} mark
  first and second per row.%
}
\label{tab:longseq}
\footnotesize
\setlength{\tabcolsep}{5pt}
\begin{tabular}{ll ccccc}
\toprule
Shots & Method & NC $\uparrow$ & Story $\uparrow$ & Char $\uparrow$ & TSIM $\uparrow$ & Pace $\uparrow$ \\
\midrule
\multirow{4}{*}{3--4}
& LTX-2                          & 4.45 & \underline{4.64} & 4.36 & \underline{0.359} & \underline{3.64} \\
& ID-LoRA                        & \underline{4.55} & 4.36 & 4.27 & 0.354 & 3.55 \\
& MOVA                           & 4.50 & 4.55 & \underline{4.45} & 0.310 & 3.09 \\
& \textbf{UnityShots~(I2V)}       & \textbf{4.98} & \textbf{4.91} & \textbf{5.00} & \textbf{0.389} & \textbf{3.73} \\
\midrule
\multirow{4}{*}{5}
& LTX-2                          & 4.43 & 4.31 & 4.65 & \underline{0.358} & 3.48 \\
& ID-LoRA                        & \underline{4.55} & \underline{4.41} & 4.65 & 0.339 & \underline{3.96} \\
& MOVA                           & 4.52 & 4.28 & \underline{4.81} & 0.287 & 4.31 \\
& \textbf{UnityShots~(I2V)}       & \textbf{4.91} & \textbf{4.87} & \textbf{4.95} & \textbf{0.378} & \textbf{4.36} \\
\midrule
\multirow{4}{*}{6+}
& LTX-2                          & 4.16 & 3.86 & 4.29 & 0.398 & 3.64 \\
& ID-LoRA                        & 4.36 & 4.14 & 4.64 & \textbf{0.449} & 3.86 \\
& MOVA                           & \underline{4.71} & \underline{4.50} & \underline{4.93} & 0.328 & \underline{4.14} \\
& \textbf{UnityShots~(I2V)}       & \textbf{5.00} & \textbf{5.00} & \textbf{5.00} & \underline{0.437} & \textbf{4.21} \\
\bottomrule
\end{tabular}
\end{table}

\subsection{Human Evaluation}
\label{sec:userstudy}

\begin{figure*}[p]
  \begin{subfigure}[t]{0.52\linewidth}
    \centering
    \includegraphics[width=\linewidth]{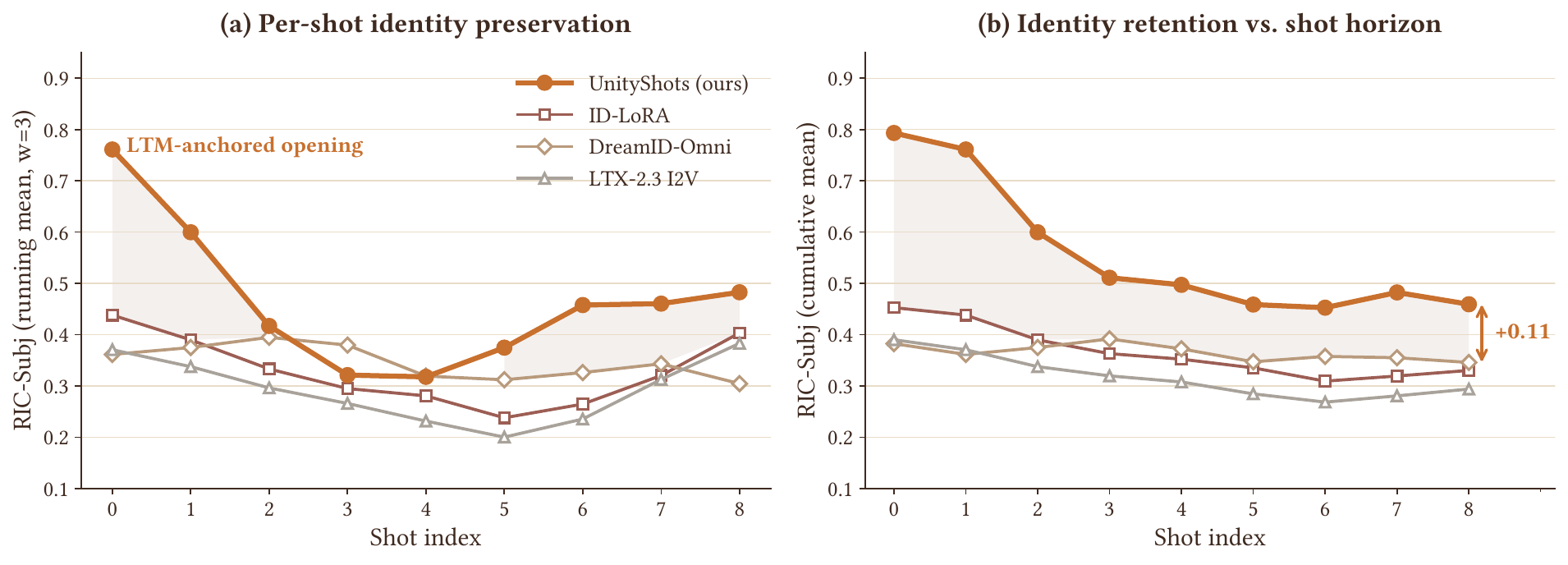}
    \caption{%
      \textbf{Per-shot identity preservation across the shot horizon.}
      (a) Per-shot NC score with a sliding-window mean: UnityShots holds its LTM-anchored
      lead across all shot indices while baselines degrade.
      (b) Cumulative-mean NC versus shot count: UnityShots dominates every prefix length
      and ends $+0.11$ above the strongest baseline.%
    }
    \label{fig:longseq}
  \end{subfigure}
  \hfill
  \begin{subfigure}[t]{0.44\linewidth}
    \centering
    \includegraphics[width=\linewidth]{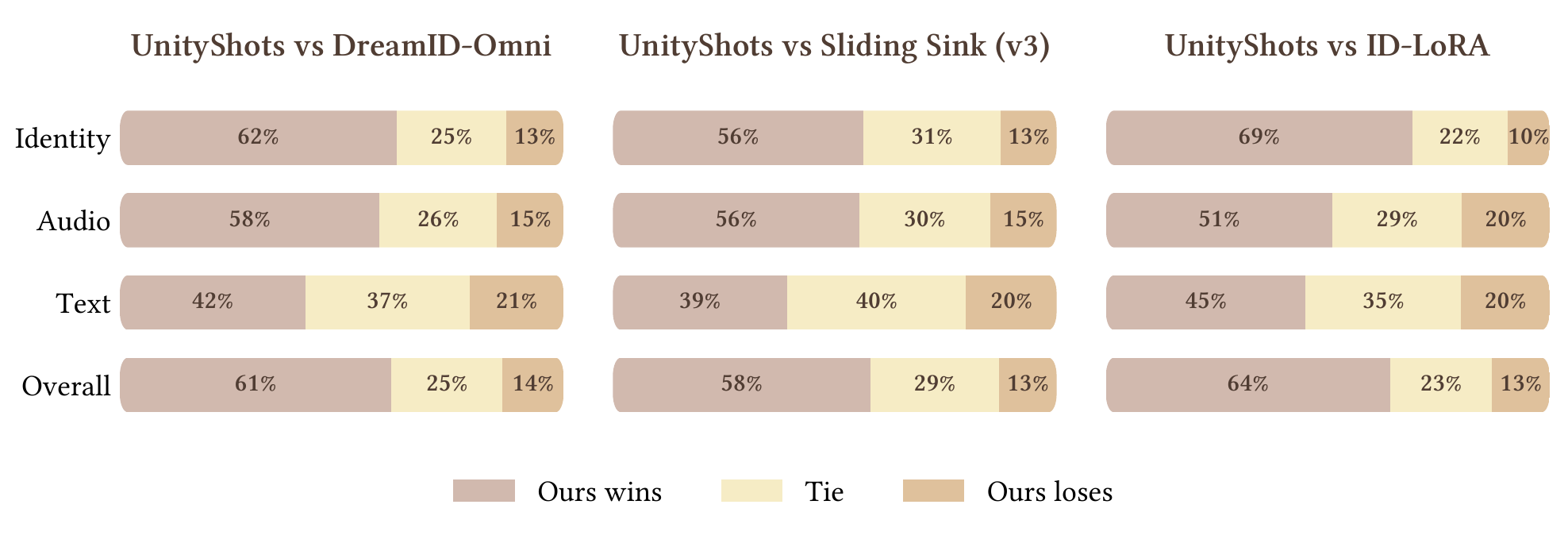}
    \caption{%
      \textbf{Pairwise human-evaluation preference rates.}
      32 participants compared 4-shot sequences from UnityShots and each competing method
      on identity consistency, audio continuity, text faithfulness, and overall quality.
      UnityShots receives majority preference on all criteria against open-source baselines.%
    }
    \label{fig:human_eval}
  \end{subfigure}
  \caption{%
    Long-sequence identity preservation (left) and human-evaluation results (right).
    See Sections~\ref{sec:longseq} and~\ref{sec:userstudy} for analysis.%
  }
  \label{fig:combined}
\end{figure*}

We conducted a user study with 32 participants using pairwise win-rate methodology
(Figure~\ref{fig:human_eval}): each participant rated 4-shot sequences on identity
consistency, audio continuity, text faithfulness, and overall quality.
UnityShots receives majority preference on identity consistency and overall quality against
all open-source baselines, with the largest audio-continuity advantage over
DreamID-Omni, which has no cross-shot audio anchor.
The preference ordering on identity consistency closely tracks the Char and NC ranking
from Table~\ref{tab:competitor}, validating that Gemini ratings capture the same
perceptual axis as human annotators.

\begin{figure*}[t]
  \centering
  \includegraphics[width=\linewidth]{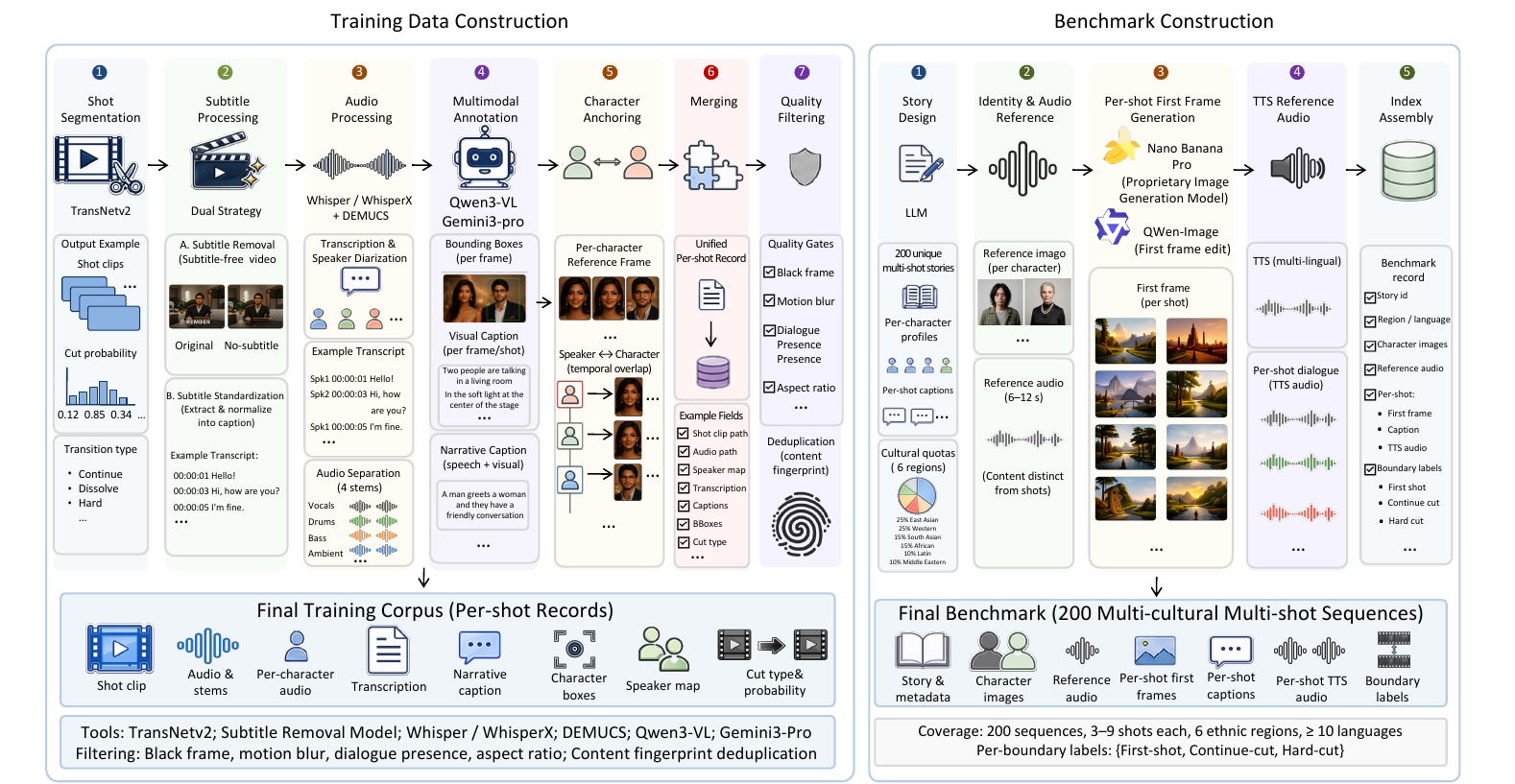}
  \caption{%
    \textbf{Training and benchmark construction pipelines.}
    \emph{Top}: the 7-stage training pipeline processes raw cinematic and music-video
    footage through shot segmentation (TransNetv2), subtitle removal, audio diarisation
    (Whisper/DEMUCS), multimodal captioning (Qwen3-VL/Qwen3-Omni~\cite{Qwen3-VL}),
    character anchoring, record merging, and quality filtering, yielding 146k annotated
    shot records.
    \emph{Bottom}: the benchmark pipeline generates the 200-sequence multi-cultural
    evaluation set through LLM story generation, reference portrait synthesis,
    per-shot first-frame generation, reference audio synthesis via multilingual TTS,
    and flat index assembly.
    Full pipeline details are in Appendix~C.%
  }
  \label{fig:data_pipe}
\end{figure*}

\section{Conclusion}
\label{sec:conclusion}

We presented \textbf{UnityShots}, a memory-driven multi-shot audio-video generation
system built on a 22B dual-stream diffusion transformer.
Two fixed-size video memory slots---long-term (opening-shot anchor) and short-term
(preceding tail)---are updated at every cut by a boundary-conditioned gate that fuses
visual and audio boundary signals.
A discrete cut-type prior learned through AdaLN gives users explicit control over
transition strength at inference, while Strata-RoPE keeps the three memory strata
on disjoint positional bands so the backbone can distinguish them without extra tokens.
Trained jointly under I2V, T2V, and R2V conditioning, UnityShots leads open-source
baselines on all cross-shot coherence metrics and matches the strongest closed-source
system on multi-shot axes across a 200-sequence multi-cultural benchmark.

\paragraph{Limitations and future work.}
Current limitations include occasional instability in subtitle rendering
across shots and visual degradation in scenes with multiple simultaneous
speakers or compositionally complex layouts; both are expected to improve
as base-model capacity grows and the companion agent system matures.
A natural next step is to leverage the agent system for iterative per-shot
frame refinement after the initial generation pass, enabling a fully automated
pipeline that detects, localizes, and corrects visual artifacts across shots
while preserving narrative continuity, and produces cinema-quality long-form
content without manual intervention.
We release model checkpoints, the multi-cultural benchmark, evaluation code,
and a companion agent system to support further work on demographically diverse
multi-shot generation.

{
    \small
    \bibliographystyle{ieeenat_fullname}
    \bibliography{main}

@String(ICASSP=	{ICASSP})

@String(AAAI = {AAAI})

@article{ho2022video,
  title={Video diffusion models},
  author={Ho, Jonathan and Salimans, Tim and Gritsenko, Alexey and Chan, William and Norouzi, Mohammad and Fleet, David J},
  journal={Advances in neural information processing systems},
  volume={35},
  pages={8633--8646},
  year={2022}
}

@article{ho2022imagen,
  title={Imagen video: High definition video generation with diffusion models},
  author={Ho, Jonathan and Chan, William and Saharia, Chitwan and Whang, Jay and Gao, Ruiqi and Gritsenko, Alexey and Kingma, Diederik P and Poole, Ben and Norouzi, Mohammad and Fleet, David J and others},
  journal={arXiv preprint arXiv:2210.02303},
  year={2022}
}

@inproceedings{peebles2023scalable,
  title={Scalable diffusion models with transformers},
  author={Peebles, William and Xie, Saining},
  booktitle={Proceedings of the IEEE/CVF international conference on computer vision},
  pages={4195--4205},
  year={2023}
}

@article{brooks2024video,
  title={Video generation models as world simulators},
  author={Brooks, Tim and Peebles, Bill and Holmes, Connor and DePue, Will and Guo, Yufei and Jing, Leo and Schnurr, David and Taylor, Joe and Luhman, Troy and Luhman, Eric and others},
  journal={OpenAI Blog},
  volume={1},
  number={8},
  pages={1},
  year={2024}
}

@article{kong2024hunyuanvideo,
  title={Hunyuanvideo: A systematic framework for large video generative models},
  author={Kong, Weijie and Tian, Qi and Zhang, Zijian and Min, Rox and Dai, Zuozhuo and Zhou, Jin and Xiong, Jiangfeng and Li, Xin and Wu, Bo and Zhang, Jianwei and others},
  journal={arXiv preprint arXiv:2412.03603},
  year={2024}
}

@article{wan2025wan,
  title={Wan: Open and advanced large-scale video generative models},
  author={Wan, Team and Wang, Ang and Ai, Baole and Wen, Bin and Mao, Chaojie and Xie, Chen-Wei and Chen, Di and Yu, Feiwu and Zhao, Haiming and Yang, Jianxiao and others},
  journal={arXiv preprint arXiv:2503.20314},
  year={2025}
}

@article{zhang2025storymem,
  title={StoryMem: Multi-shot Long Video Storytelling with Memory},
  author={Zhang, Kaiwen and Jiang, Liming and Wang, Angtian and Fang, Jacob Zhiyuan and Zhi, Tiancheng and Yan, Qing and Kang, Hao and Lu, Xin and Pan, Xingang},
  journal={arXiv preprint arXiv:2512.19539},
  year={2025}
}

@article{wang2025multishotmaster,
  title={Multishotmaster: A controllable multi-shot video generation framework},
  author={Wang, Qinghe and Shi, Xiaoyu and Li, Baolu and Bian, Weikang and Liu, Quande and Lu, Huchuan and Wang, Xintao and Wan, Pengfei and Gai, Kun and Jia, Xu},
  journal={arXiv preprint arXiv:2512.03041},
  year={2025}
}

@article{meng2025holocine,
  title={Holocine: Holistic generation of cinematic multi-shot long video narratives},
  author={Meng, Yihao and Ouyang, Hao and Yu, Yue and Wang, Qiuyu and Wang, Wen and Cheng, Ka Leong and Wang, Hanlin and Li, Yixuan and Chen, Cheng and Zeng, Yanhong and others},
  journal={arXiv preprint arXiv:2510.20822},
  year={2025}
}

@article{guo2026dreamid,
  title={DreamID-Omni: Unified Framework for Controllable Human-Centric Audio-Video Generation},
  author={Guo, Xu and Ye, Fulong and Sun, Qichao and Chen, Liyang and Li, Bingchuan and Zhang, Pengze and Liu, Jiawei and Zhao, Songtao and He, Qian and Hou, Xiangwang},
  journal={arXiv preprint arXiv:2602.12160},
  year={2026}
}

@article{soucek2020transnet,
  title={TransNet V2: An effective deep network architecture for fast shot transition detection. CoRR abs/2008.04838 (2020)},
  author={Soucek, Tom{\'a}s and Lokoc, Jakub},
  journal={arXiv preprint arXiv:2008.04838},
  year={2020}
}

@article{foscarin2024beat,
  title={Beat this! accurate beat tracking without dbn postprocessing},
  author={Foscarin, Francesco and Schl{\"u}ter, Jan and Widmer, Gerhard},
  journal={arXiv preprint arXiv:2407.21658},
  year={2024}
}

@inproceedings{wang2024internvid,
  title={Internvid: A large-scale video-text dataset for multimodal understanding and generation},
  author={Wang, Yi and He, Yinan and Li, Yizhuo and Li, Kunchang and Yu, Jiashuo and Ma, Xin and Li, Xinhao and Chen, Guo and Chen, Xinyuan and Wang, Yaohui and others},
  booktitle={International Conference on Learning Representations},
  volume={2024},
  pages={42055--42079},
  year={2024}
}

@inproceedings{huang2024vbench,
  title={Vbench: Comprehensive benchmark suite for video generative models},
  author={Huang, Ziqi and He, Yinan and Yu, Jiashuo and Zhang, Fan and Si, Chenyang and Jiang, Yuming and Zhang, Yuanhan and Wu, Tianxing and Jin, Qingyang and Chanpaisit, Nattapol and others},
  booktitle={Proceedings of the IEEE/CVF Conference on Computer Vision and Pattern Recognition},
  pages={21807--21818},
  year={2024}
}

@inproceedings{rajbhandari2020zero,
  title={Zero: Memory optimizations toward training trillion parameter models},
  author={Rajbhandari, Samyam and Rasley, Jeff and Ruwase, Olatunji and He, Yuxiong},
  booktitle={SC20: international conference for high performance computing, networking, storage and analysis},
  pages={1--16},
  year={2020},
  organization={IEEE}
}

@article{zhou2024storydiffusion,
  title={Storydiffusion: Consistent self-attention for long-range image and video generation},
  author={Zhou, Yupeng and Zhou, Daquan and Cheng, Ming-Ming and Feng, Jiashi and Hou, Qibin},
  journal={Advances in Neural Information Processing Systems},
  volume={37},
  pages={110315--110340},
  year={2024}
}

@article{team2026mova,
  title={Mova: Towards scalable and synchronized video-audio generation},
  author={Team, OpenMOSS and Yu, Donghua and Chen, Mingshu and Chen, Qi and Luo, Qi and Wu, Qianyi and Cheng, Qinyuan and Li, Ruixiao and Liang, Tianyi and Zhang, Wenbo and others},
  journal={arXiv preprint arXiv:2602.08794},
  year={2026}
}

@article{dahan2026id,
  title={ID-LoRA: Identity-Driven Audio-Video Personalization with In-Context LoRA},
  author={Dahan, Aviad and Yanuka, Moran and Kraicer, Noa and Wolf, Lior and Giryes, Raja},
  journal={arXiv preprint arXiv:2603.10256},
  year={2026}
}

@article{low2025ovi,
  title={Ovi: Twin backbone cross-modal fusion for audio-video generation},
  author={Low, Chetwin and Wang, Weimin and Katyal, Calder},
  journal={arXiv preprint arXiv:2510.01284},
  year={2025}
}

@article{hacohen2026ltx,
  title={LTX-2: Efficient Joint Audio-Visual Foundation Model},
  author={HaCohen, Yoav and Brazowski, Benny and Chiprut, Nisan and Bitterman, Yaki and Kvochko, Andrew and Berkowitz, Avishai and Shalem, Daniel and Lifschitz, Daphna and Moshe, Dudu and Porat, Eitan and others},
  journal={arXiv preprint arXiv:2601.03233},
  year={2026}
}

@article{ho2020denoising,
  title={Denoising diffusion probabilistic models},
  author={Ho, Jonathan and Jain, Ajay and Abbeel, Pieter},
  journal={Advances in neural information processing systems},
  volume={33},
  pages={6840--6851},
  year={2020}
}

@article{song2020denoising,
  title={Denoising diffusion implicit models},
  author={Song, Jiaming and Meng, Chenlin and Ermon, Stefano},
  journal={arXiv preprint arXiv:2010.02502},
  year={2020}
}

@inproceedings{rombach2022high,
  title={High-resolution image synthesis with latent diffusion models},
  author={Rombach, Robin and Blattmann, Andreas and Lorenz, Dominik and Esser, Patrick and Ommer, Bj{\"o}rn},
  booktitle={Proceedings of the IEEE/CVF conference on computer vision and pattern recognition},
  pages={10684--10695},
  year={2022}
}

@article{vaswani2017attention,
  title={Attention is all you need},
  author={Vaswani, Ashish and Shazeer, Noam and Parmar, Niki and Uszkoreit, Jakob and Jones, Llion and Gomez, Aidan N and Kaiser, {\L}ukasz and Polosukhin, Illia},
  journal={Advances in neural information processing systems},
  volume={30},
  year={2017}
}

@article{su2024roformer,
  title={Roformer: Enhanced transformer with rotary position embedding},
  author={Su, Jianlin and Ahmed, Murtadha and Lu, Yu and Pan, Shengfeng and Bo, Wen and Liu, Yunfeng},
  journal={Neurocomputing},
  volume={568},
  pages={127063},
  year={2024},
  publisher={Elsevier}
}

@article{blattmann2023stable,
  title={Stable video diffusion: Scaling latent video diffusion models to large datasets},
  author={Blattmann, Andreas and Dockhorn, Tim and Kulal, Sumith and Mendelevitch, Daniel and Kilian, Maciej and Lorenz, Dominik and Levi, Yam and English, Zion and Voleti, Vikram and Letts, Adam and others},
  journal={arXiv preprint arXiv:2311.15127},
  year={2023}
}

@inproceedings{yang2025cogvideox,
  title={Cogvideox: Text-to-video diffusion models with an expert transformer},
  author={Yang, Zhuoyi and Teng, Jiayan and Zheng, Wendi and Ding, Ming and Huang, Shiyu and Xu, Jiazheng and Yang, Yuanming and Hong, Wenyi and Zhang, Xiaohan and Feng, Guanyu and others},
  booktitle={International Conference on Learning Representations},
  volume={2025},
  pages={83048--83077},
  year={2025}
}

@article{polyak2024movie,
  title={Movie gen: A cast of media foundation models},
  author={Polyak, Adam and Zohar, Amit and Brown, Andrew and Tjandra, Andros and Sinha, Animesh and Lee, Ann and Vyas, Apoorv and Shi, Bowen and Ma, Chih-Yao and Chuang, Ching-Yao and others},
  journal={arXiv preprint arXiv:2410.13720},
  year={2024}
}

@article{zheng2024open,
  title={Open-sora: Democratizing efficient video production for all},
  author={Zheng, Zangwei and Peng, Xiangyu and Yang, Tianji and Shen, Chenhui and Li, Shenggui and Liu, Hongxin and Zhou, Yukun and Li, Tianyi and You, Yang},
  journal={arXiv preprint arXiv:2412.20404},
  year={2024}
}

@article{liu2024audioldm,
  title={Audioldm 2: Learning holistic audio generation with self-supervised pretraining},
  author={Liu, Haohe and Yuan, Yi and Liu, Xubo and Mei, Xinhao and Kong, Qiuqiang and Tian, Qiao and Wang, Yuping and Wang, Wenwu and Wang, Yuxuan and Plumbley, Mark D},
  journal={IEEE/ACM Transactions on Audio, Speech, and Language Processing},
  volume={32},
  pages={2871--2883},
  year={2024},
  publisher={IEEE}
}

@inproceedings{elizalde2023clap,
  title={Clap learning audio concepts from natural language supervision},
  author={Elizalde, Benjamin and Deshmukh, Soham and Al Ismail, Mahmoud and Wang, Huaming},
  booktitle={ICASSP 2023-2023 IEEE International Conference on Acoustics, Speech and Signal Processing (ICASSP)},
  pages={1--5},
  year={2023},
  organization={IEEE}
}

@article{zhang2026foleycrafter,
  title={Foleycrafter: Bring silent videos to life with lifelike and synchronized sounds},
  author={Zhang, Yiming and Gu, Yicheng and Zeng, Yanhong and Xing, Zhening and Wang, Yuancheng and Wu, Zhizheng and Liu, Bin and Chen, Kai},
  journal={International Journal of Computer Vision},
  volume={134},
  number={1},
  pages={46},
  year={2026},
  publisher={Springer}
}

@article{tjandra2025meta,
  title={Meta audiobox aesthetics: Unified automatic quality assessment for speech, music, and sound},
  author={Tjandra, Andros and Wu, Yi-Chiao and Guo, Baishan and Hoffman, John and Ellis, Brian and Vyas, Apoorv and Shi, Bowen and Chen, Sanyuan and Le, Matt and Zacharov, Nick and others},
  journal={arXiv preprint arXiv:2502.05139},
  year={2025}
}

@article{oquab2023dinov2,
  title={Dinov2: Learning robust visual features without supervision},
  author={Oquab, Maxime and Darcet, Timoth{\'e}e and Moutakanni, Th{\'e}o and Vo, Huy and Szafraniec, Marc and Khalidov, Vasil and Fernandez, Pierre and Haziza, Daniel and Massa, Francisco and El-Nouby, Alaaeldin and others},
  journal={arXiv preprint arXiv:2304.07193},
  year={2023}
}

@article{ye2023ip,
  title={Ip-adapter: Text compatible image prompt adapter for text-to-image diffusion models},
  author={Ye, Hu and Zhang, Jun and Liu, Sibo and Han, Xiao and Yang, Wei},
  journal={arXiv preprint arXiv:2308.06721},
  year={2023}
}

@article{zhang2026frame,
  title={Frame context packing and drift prevention in next-frame-prediction video diffusion models},
  author={Zhang, Lvmin and Cai, Shengqu and Li, Muyang and Wetzstein, Gordon and Agrawala, Maneesh},
  journal={Advances in Neural Information Processing Systems},
  volume={38},
  pages={30546--30566},
  year={2026}
}

@misc{wu2025qwenimagetechnicalreport,
      title={Qwen-Image Technical Report}, 
      author={Chenfei Wu and Jiahao Li and Jingren Zhou and Junyang Lin and Kaiyuan Gao and Kun Yan and Sheng-ming Yin and Shuai Bai and Xiao Xu and Yilei Chen and Yuxiang Chen and Zecheng Tang and Zekai Zhang and Zhengyi Wang and An Yang and Bowen Yu and Chen Cheng and Dayiheng Liu and Deqing Li and Hang Zhang and Hao Meng and Hu Wei and Jingyuan Ni and Kai Chen and Kuan Cao and Liang Peng and Lin Qu and Minggang Wu and Peng Wang and Shuting Yu and Tingkun Wen and Wensen Feng and Xiaoxiao Xu and Yi Wang and Yichang Zhang and Yongqiang Zhu and Yujia Wu and Yuxuan Cai and Zenan Liu},
      year={2025},
      eprint={2508.02324},
      archivePrefix={arXiv},
      primaryClass={cs.CV},
      url={https://arxiv.org/abs/2508.02324}, 
}

@article{huang2024iclora,
  title={In-context {LoRA} for diffusion transformers},
  author={Huang, Lianghua and Wang, Wei and Wu, Zhi-Fan and Shi, Yupeng and Dou, Huanzhang and Liang, Chen and Feng, Yutong and Liu, Yu and Zhou, Jingren},
  journal={arXiv preprint arXiv:2410.23775},
  year={2024}
}

@article{guo2025lct,
  title={Long context tuning for video generation},
  author={Guo, Yuwei and Yang, Ceyuan and Yang, Ziyan and Ma, Zhibei and Lin, Zhijie and Yang, Zhenheng and Lin, Dahua and Jiang, Lu},
  journal={arXiv preprint arXiv:2503.10589},
  year={2025}
}

@inproceedings{kara2025shotadapter,
  title={Shotadapter: Text-to-multi-shot video generation with diffusion models},
  author={Kara, Ozgur and Singh, Krishna Kumar and Liu, Feng and Ceylan, Duygu and Rehg, James M and Hinz, Tobias},
  booktitle={Proceedings of the Computer Vision and Pattern Recognition Conference},
  pages={28405--28415},
  year={2025}
}

@article{xiao2025captaincinema,
  title={Captain cinema: Towards short movie generation},
  author={Xiao, Junfei and Yang, Ceyuan and Zhang, Lvmin and Cai, Shengqu and Zhao, Yang and Guo, Yuwei and Wetzstein, Gordon and Agrawala, Maneesh and Yuille, Alan L and Jiang, Lu},
  journal={arXiv preprint arXiv:2507.18634},
  year={2025}
}

@inproceedings{wang2025cinemaster,
  title={Cinemaster: A 3d-aware and controllable framework for cinematic text-to-video generation},
  author={Wang, Qinghe and Luo, Yawen and Shi, Xiaoyu and Jia, Xu and Lu, Huchuan and Xue, Tianfan and Wang, Xintao and Wan, Pengfei and Zhang, Di and Gai, Kun},
  booktitle={Proceedings of the Special Interest Group on Computer Graphics and Interactive Techniques Conference Conference Papers},
  pages={1--10},
  year={2025}
}

@inproceedings{xing2025motioncanvas,
  title={Motioncanvas: Cinematic shot design with controllable image-to-video generation},
  author={Xing, Jinbo and Mai, Long and Ham, Cusuh and Huang, Jiahui and Mahapatra, Aniruddha and Fu, Chi-Wing and Wong, Tien-Tsin and Liu, Feng},
  booktitle={Proceedings of the Special Interest Group on Computer Graphics and Interactive Techniques Conference Conference Papers},
  pages={1--11},
  year={2025}
}

@article{he2025cut2next,
  title={Cut2next: Generating next shot via in-context tuning},
  author={He, Jingwen and Liu, Hongbo and Li, Jiajun and Huang, Ziqi and Qiao, Yu and Ouyang, Wanli and Liu, Ziwei},
  journal={arXiv preprint arXiv:2508.08244},
  year={2025}
}

@article{wu2025cinetrans,
  title={Cinetrans: Learning to generate videos with cinematic transitions via masked diffusion models},
  author={Wu, Xiaoxue and Gao, Bingjie and Qiao, Yu and Wang, Yaohui and Chen, Xinyuan},
  journal={arXiv preprint arXiv:2508.11484},
  year={2025}
}

@article{luo2026shotstream,
  title={ShotStream: Streaming Multi-Shot Video Generation for Interactive Storytelling},
  author={Luo, Yawen and Shi, Xiaoyu and Zhuang, Junhao and Chen, Yutian and Liu, Quande and Wang, Xintao and Wan, Pengfei and Xue, Tianfan},
  journal={arXiv preprint arXiv:2603.25746},
  year={2026}
}

@article{zhou2026videomemory,
  title={VideoMemory: Toward Consistent Video Generation via Memory Integration},
  author={Zhou, Jinsong and Du, Yihua and Xu, Xinli and Wang, Luozhou and Zhuang, Zijie and Zhang, Yehang and Li, Shuaibo and Hu, Xiaojun and Su, Bolan and Chen, Ying-cong},
  journal={arXiv preprint arXiv:2601.03655},
  year={2026}
}

@article{liang2025univa,
  title={UniVA: Universal Video Agent towards Open-Source Next-Generation Video Generalist},
  author={Liang, Zhengyang and Zhang, Daoan and Zhou, Huichi and Huang, Rui and Li, Bobo and Zhang, Yuechen and Wu, Shengqiong and Wang, Xiaohan and Luo, Jiebo and Liao, Lizi and others},
  journal={arXiv preprint arXiv:2511.08521},
  year={2025}
}

@article{li2026direct,
  title={DIRECT: Video Mashup Creation via Hierarchical Multi-Agent Planning and Intent-Guided Editing},
  author={Li, Ke and Li, Maoliang and Chen, Jialiang and Chen, Jiayu and Zheng, Zihao and Wang, Shaoqi and Chen, Xiang},
  journal={arXiv preprint arXiv:2604.04875},
  year={2026}
}

@misc{kling2026,
    title={Kling AI},
    author={{Kuaishou Technology}},
    howpublished={\url{https://klingai.com/}},
    year={2026}
}

@misc{seedance2025,
    title={Seedance 2.0},
    author={{ByteDance}},
    howpublished={\url{https://seed.bytedance.com/en/seed2}},
    year={2026}
}

@article{bain2023whisperx,
  title={Whisperx: Time-accurate speech transcription of long-form audio},
  author={Bain, Max and Huh, Jaesung and Han, Tengda and Zisserman, Andrew},
  journal={arXiv preprint arXiv:2303.00747},
  year={2023}
}

@article{defossez2021hybrid,
  title={Hybrid spectrogram and waveform source separation},
  author={D{\'e}fossez, Alexandre},
  journal={arXiv preprint arXiv:2111.03600},
  year={2021}
}

@article{Qwen3-VL,
      title={Qwen3-VL Technical Report}, 
      author={Shuai Bai and Yuxuan Cai and Ruizhe Chen and Keqin Chen and Xionghui Chen and Zesen Cheng and Lianghao Deng and Wei Ding and Chang Gao and Chunjiang Ge and Wenbin Ge and Zhifang Guo and Qidong Huang and Jie Huang and Fei Huang and Binyuan Hui and Shutong Jiang and Zhaohai Li and Mingsheng Li and Mei Li and Kaixin Li and Zicheng Lin and Junyang Lin and Xuejing Liu and Jiawei Liu and Chenglong Liu and Yang Liu and Dayiheng Liu and Shixuan Liu and Dunjie Lu and Ruilin Luo and Chenxu Lv and Rui Men and Lingchen Meng and Xuancheng Ren and Xingzhang Ren and Sibo Song and Yuchong Sun and Jun Tang and Jianhong Tu and Jianqiang Wan and Peng Wang and Pengfei Wang and Qiuyue Wang and Yuxuan Wang and Tianbao Xie and Yiheng Xu and Haiyang Xu and Jin Xu and Zhibo Yang and Mingkun Yang and Jianxin Yang and An Yang and Bowen Yu and Fei Zhang and Hang Zhang and Xi Zhang and Bo Zheng and Humen Zhong and Jingren Zhou and Fan Zhou and Jing Zhou and Yuanzhi Zhu and Ke Zhu},
	  journal={arXiv preprint arXiv:2511.21631},
      year={2025}
}

@inproceedings{luo2026filmweaver,
  title={Filmweaver: Weaving consistent multi-shot videos with cache-guided autoregressive diffusion},
  author={Luo, Xiangyang and Li, Qingyu and Liu, Xiaokun and Qin, Wenyu and Yang, Miao and Wang, Meng and Wan, Pengfei and Zhang, Di and Gai, Kun and Huang, Shao-Lun},
  booktitle={Proceedings of the AAAI Conference on Artificial Intelligence},
  volume={40},
  number={9},
  pages={7689--7697},
  year={2026}
}

@article{an2025onestory,
  title={Onestory: Coherent multi-shot video generation with adaptive memory},
  author={An, Zhaochong and Jia, Menglin and Qiu, Haonan and Zhou, Zijian and Huang, Xiaoke and Liu, Zhiheng and Ren, Weiming and Kahatapitiya, Kumara and Liu, Ding and He, Sen and others},
  journal={arXiv preprint arXiv:2512.07802},
  year={2025}
}

@article{zhang2025stage,
  title={STAGE: Storyboard-Anchored Generation for Cinematic Multi-shot Narrative},
  author={Zhang, Peixuan and Jia, Zijian and Liu, Kaiqi and Weng, Shuchen and Li, Si and Shi, Boxin},
  journal={arXiv preprint arXiv:2512.12372},
  year={2025}
}
}

\clearpage
\clearpage
\setcounter{page}{1}

\maketitlesupplementary

\appendix

\setcounter{figure}{0}
\setcounter{table}{0}

\section*{Appendix}
\addcontentsline{toc}{section}{Appendix}

\subsection*{A. Training and Architecture Details}
\label{sec:supp-config}

Table~\ref{tab:supp-config} lists the key hyperparameters for Stage~$2$ multi-shot
fine-tuning. Stage~$1$ (identity foundation) uses the same base model, optimiser,
and learning rate, training on single-clip identity data.

\begin{table}[h]
\centering
\small
\caption{Stage~$2$ training hyperparameters.}
\label{tab:supp-config}
\setlength{\tabcolsep}{6pt}
\begin{tabularx}{\columnwidth}{@{}lX@{}}
\toprule
\textbf{Parameter} & \textbf{Value} \\
\midrule
Base model & LTX-2.3 22B (\texttt{ltx-2.3-22b-dev}) \\
Initialization & Stage~$1$ checkpoint \\
Optimizer & AdamW, $\beta_1{=}0.9$, $\beta_2{=}0.999$ \\
Learning rate & $1\!\times\!10^{-5}$; gate MLP $5\!\times\!10^{-5}$ \\
Gradient clipping & $1.0$ \\
Precision & BFloat16, DeepSpeed \\
Batch size & $1$ per GPU, gradient accumulation $1$ \\
Max latent frames & $96$ ($\approx\!30$\,s at $25$\,fps, $8\!\times$ compression) \\
\midrule
Video LTM slot & $2$ latent frames \\
Video STM slot ($P_v$ frames) & $4$ latent frames ($\approx\!1$\,s) \\
Audio reference anchor & $1$ speaker-identity token from $\mathbf{A}^{\mathrm{ref}}$ \\
Video RoPE ceiling ($f_{\mathrm{lim}}$) & $128$ (current shot: $[0, T_v{-}1]$) \\
Video STM band & $[f_{\mathrm{lim}}{-}P_v,\ f_{\mathrm{lim}}{-}1]$ \\
Video LTM band & $[f_{\mathrm{lim}}{-}2P_v,\ f_{\mathrm{lim}}{-}P_v{-}1]$ \\
\midrule
Cut-type prior $\tau_N$ & $\{\textsc{first}, \textsc{continue}, \textsc{hard}\} \to \{0,\,0.4,\,1.0\}$, fused with timestep embedding via AdaLN \\
Boundary weights (cinematic) & $\alpha{=}0.5,\ \beta{=}0.3,\ \gamma{=}0.0$ \\
Boundary weights (music video) & $\alpha{=}0.5,\ \beta{=}0.3,\ \gamma{=}1.0$ \\
STM coefficient & $g_{\mathrm{stm}}(b) = 0.3 + 0.7\,b$ (empirically determined) \\
LTM coefficient & $g_{\mathrm{ltm}}(b) = 0.1 + 0.6\,b$ (empirically determined) \\
LTM recurrence weight & $z_N = \mathrm{clamp}(b_N, 0, 1)$ \\
Content-aware MLP & $2$-layer, $<\!50$K parameters; warm-up clamp $[0.5, 1.0]$ for $500$ steps \\
$k$-shot training chunks & variable $k\!\in\![3, 9]$ with bucketed-length sampling \\
Mixed conditioning ($p_{\mathrm{i2v}},p_{\mathrm{t2v}},p_{\mathrm{r2v}}$) & $0.5,\,0.3,\,0.2$ \\
No-subtitle probability & $0.3$ \\
Reference-audio dropout & $0.1$ \\
\bottomrule
\end{tabularx}
\end{table}

\subsection*{B. Benchmark Summary}
\label{sec:supp-bench}

The evaluation benchmark contains $200$ multi-cultural multi-shot sequences with
$3$--$9$ shots each, covering six ethnic regions and ten or more languages.
Cultural quotas ensure coverage: $25\%$ East Asian, $25\%$ Western, $15\%$ South
Asian, $15\%$ African, $10\%$ Latin, $10\%$ Middle Eastern.
Each sequence is annotated with reference identity images, reference audio
($6$--$12$\,s, content distinct from shot dialogue), per-shot first frames
generated by an image-generation model conditioned on each character portrait and per-shot
caption, per-shot captions, and per-boundary transition-type labels
(\textsc{first-shot}, \textsc{continue-cut}, \textsc{hard-cut}).
The full construction pipeline---story design, identity reference, per-shot first
frame, TTS reference audio, and index assembly---is described in Appendix~C
together with the training-data pipeline, since the two share the same tooling.

\subsection*{C. Training Data and Benchmark Construction}
\label{sec:supp-data}

This appendix details the pipelines used to construct (i) the multi-shot training
corpus and (ii) the multi-cultural evaluation benchmark.
A summary figure of the overall workflow will accompany the camera-ready version.

\paragraph{Training data construction.}
Multi-shot training records are produced by a seven-stage pipeline.
\textbf{(1) Shot segmentation.}
TransNetv2 segments raw video into individual shot clips and
returns a per-boundary visual cut probability ~\cite{soucek2020transnet}  and an inferred transition type
(\textsc{continue}, \textsc{dissolve}, \textsc{hard}) used to populate the cut-type
prior in the boundary-conditioned gating module.
\textbf{(2) Subtitle handling.}
Each shot is annotated with its burnt-in subtitle text, if any.
During training, $20\%$ of samples randomly apply a subtitle-removal model to produce
a subtitle-free variant, enabling the model to generate clean footage or captioned
footage depending on the text prompt without distribution mismatch.
\textbf{(3) Audio processing.}
We extract the soundtrack and run WhisperX~\cite{bain2023whisperx} for transcription and
per-speaker diarisation.
DEMUCS~\cite{defossez2021hybrid} separates vocals, drums, bass, and ambient stems for downstream
conditioning.
\textbf{(4) Multimodal annotation.}
Qwen3-VL~\cite{Qwen3-VL} provides initial per-frame character bounding-box detection
and coarse visual captions.
Gemini3-Pro then performs fine-grained narrative captioning that fuses speech
transcripts with visual events.
\textbf{(5) Character anchoring.}
For each shot we extract a per-character reference frame using an image-generation
model for identity-preserving portrait extraction, and link acoustic speaker labels from
WhisperX to visual bounding boxes via temporal-overlap matching, producing a
per-shot speaker-to-character map.
\textbf{(6) Merging.}
Per-stage outputs are consolidated into a single per-shot record under hierarchical
priority handling so that partial annotations do not corrupt downstream training.
\textbf{(7) Quality filtering.}
Black-frame, motion-blur, dialogue-presence, and aspect-ratio gates remove unusable
shots, and content-fingerprint deduplication enforces sequence-level uniqueness.
Each surviving shot record carries the clip path, audio path, per-character isolated
audio paths, transcription, narrative caption, character bounding-box metadata, and
boundary annotations.
We do not disclose source titles; only the tooling above is publicly documented.

\paragraph{Benchmark construction.}
The $200$-sequence multi-cultural benchmark is built by a five-stage pipeline.
\textbf{(1) Story design.}
Gemini3-Pro generates $200$ unique multi-shot story outlines with per-character
descriptions and per-shot captions, balanced across six ethnic regions ($25\%$ East
Asian, $25\%$ Western, $15\%$ South Asian, $15\%$ African, $10\%$ Latin, $10\%$
Middle Eastern) and ten or more languages, with controlled coverage of
dialogue-heavy and interview-format narratives to stress audio continuity.
Stories are deduplicated across the full set to prevent theme, conflict, or character
repetition.
\textbf{(2) Identity reference.}
An image-generation model synthesises a photorealistic portrait per character,
conditioned on the character's demographic description.
\textbf{(3) Per-shot first frame.}
The same image-generation model generates a per-shot first frame conditioned on the
character portrait and the per-shot caption, ensuring visual consistency between the
identity reference and the first frame provided to I2V evaluation.
\textbf{(4) Reference audio.}
ElevenLabs \texttt{eleven\_multilingual\_v2} TTS synthesises a $6$--$12$\,s reference
clip per character; the reference content is deliberately chosen to differ from the
shot dialogue to prevent acoustic leakage between reference and generation target.
\textbf{(5) Index assembly.}
A flat \texttt{index.json} maps each sequence ID to its reference portraits, audio
clips, per-shot first frames, and per-shot captions.
The benchmark and all per-stage scripts are released alongside the model.

\subsection*{D. Extended Comparison with Published Multi-Shot Methods}
\label{sec:supp-extended}

Table~\ref{tab:supp-public-bench} extends the main-paper evaluation to additional
published T2V multi-shot methods on our $200$-sequence multi-cultural benchmark,
using the same evaluation protocol as Table~1 of the main paper.
StoryMem~\cite{zhang2025storymem} represents the shot-by-shot family;
MultiShotMaster~\cite{wang2025multishotmaster} and
HoloCine~\cite{meng2025holocine} represent the end-to-end family.
All methods are evaluated in T2V mode to remove first-frame distribution confounds.

\begin{table}[h]
\centering
\caption{%
  Extended T2V comparison with published multi-shot methods on the
  $200$-sequence multi-cultural benchmark.
  $\uparrow$ higher is better.
  \textbf{Bold} and \underline{underline} mark first and second per column.
}
\label{tab:supp-public-bench}
\setlength{\tabcolsep}{3pt}
\resizebox{\columnwidth}{!}{%
\begin{tabular}{lccccc}
\toprule
Method & TA $\uparrow$ & TSIM $\uparrow$ & NC $\uparrow$ & Story $\uparrow$ & Char $\uparrow$ \\
\midrule
StoryMem~\cite{zhang2025storymem}              & 17.31          & \underline{0.368} & 3.42          & 3.18          & 3.72 \\
MultiShotMaster~\cite{wang2025multishotmaster} & \underline{18.47} & 0.346        & \underline{3.65} & \underline{3.41} & \underline{4.05} \\
HoloCine~\cite{meng2025holocine}               & 18.16          & 0.322             & 3.51          & 3.22          & 3.72 \\
\textbf{UnityShots~(T2V) (ours)}               & \textbf{19.17} & \textbf{0.451} & \textbf{4.13} & \textbf{3.83} & \textbf{4.11} \\
\bottomrule
\end{tabular}%
}
\end{table}

StoryMem maintains higher TSIM than end-to-end methods because each shot is
conditioned on the preceding clip, but trails on NC and Story as identity drift
accumulates past the conditioning window.
MultiShotMaster achieves the strongest TA among baselines from larger-scale
training, yet falls behind on NC and TSIM because its fixed context window does
not distinguish long-range from short-range context.
UnityShots eliminates this trade-off: the dual-tier memory bank gains $+0.48$
NC and $+0.42$ Story over MultiShotMaster while also leading on TSIM by $+0.083$
over StoryMem.

\subsection*{E. Baseline Evaluation Protocol and Standard Benchmark Results}
\label{sec:supp-repro}

\paragraph{Evaluation setup.}
MOVA~\cite{team2026mova}, LTX-2~\cite{hacohen2026ltx}, ID-LoRA~\cite{dahan2026id}, OVI~\cite{low2025ovi} and DreamID-Omni~\cite{guo2026dreamid} were evaluated on the $200$-sequence multi-cultural
benchmark using official checkpoints.
For open-source baselines with available checkpoints, we run official checkpoints with the recommended inference settings and use DDIM inference with 50 steps when applicable. For closed-source or service-only systems, we collect outputs through their official interfaces using the same benchmark prompts, reference inputs, and shot-level protocol. All metrics reported in the main paper are computed by us using the same evaluation scripts, rather than copied from prior publications.

\paragraph{VBench comparison.}
Table~\ref{tab:supp-vbench} reports standard VBench~\cite{huang2024vbench}
metrics for the methods we evaluated directly, with all shots concatenated at
inferred boundaries for per-video scoring.
Subject Consistency (Subj) is the VBench metric most directly tied to
cross-shot memory: UnityShots leads by $+0.006$--$+0.014$ over baselines,
reflecting the LTM slot anchoring opening-shot identity through every denoising
pass.
Background Consistency (BG) and Motion Smoothness (MoS) also improve, since
the memory conditioning prevents abrupt scene mode changes at cuts.
Aesthetic Quality (AES) is on par with LTX-2 and ID-LoRA; the slight gap
versus MOVA is consistent with the memory regularisation mildly reducing
per-frame perceptual variety in exchange for cross-shot stability.

\begin{table}[h]
\centering
\caption{%
  VBench evaluation on our $200$-sequence benchmark (shots concatenated).
  Subj: Subject Consistency; BG: Background Consistency;
  MoS: Motion Smoothness; AES: Aesthetic Quality; TA: Text--Video Alignment.
  $\uparrow$ higher is better. \textbf{Bold} marks best per column.%
}
\label{tab:supp-vbench}
\small
\setlength{\tabcolsep}{5pt}
\begin{tabular}{lccccc}
\toprule
Method & Subj $\uparrow$ & BG $\uparrow$ & MoS $\uparrow$ & AES $\uparrow$ & TA $\uparrow$ \\
\midrule
LTX-2~\cite{hacohen2026ltx}   & 0.953 & 0.952 & 0.977 & 0.649 & 0.282 \\
ID-LoRA~\cite{dahan2026id}     & 0.961 & 0.959 & 0.980 & 0.652 & 0.289 \\
MOVA~\cite{team2026mova}       & 0.957 & 0.955 & 0.978 & \textbf{0.656} & 0.287 \\
\textbf{UnityShots~(I2V)}      & \textbf{0.967} & \textbf{0.963} & \textbf{0.982} & 0.652 & \textbf{0.295} \\
\bottomrule
\end{tabular}
\end{table}

\subsection*{F. Pseudocode for Memory Update}
\label{sec:supp-algo}

Algorithms~\ref{alg:boundary_gate} and~\ref{alg:hier_mem} give complete pseudocode for
the boundary-aware gate and the full inference loop respectively.

\begin{algorithm}[h]
\caption{Per-Shot Memory Gate Application}
\label{alg:boundary_gate}
\begin{algorithmic}[1]
\Require Prev-shot latent $\hat{\mathbf{V}}^{N-1}$; boundary strength $b \in [0,1]$;
         LTM slot $\mathbf{L}^N$; reference audio $\mathbf{A}^{\mathrm{ref}}$;
         noisy latent $\mathbf{V}^N_t$
\State Extract tail: $\mathbf{S}^N \leftarrow \hat{\mathbf{V}}^{N-1}_{:P_v}$
\State Boundary-conditioned coefficients: $g_{\mathrm{stm}} \leftarrow 0.3 + 0.7b$;\; $g_{\mathrm{ltm}} \leftarrow 0.1 + 0.6b$
\State Content-aware refinement: $\rho \leftarrow \sigma(\phi(\mathrm{pool}(\hat{\mathbf{V}}^{N-1}),\, \mathrm{pool}(\mathbf{V}^N_t),\, \mathbf{e}_\tau))$
\State Scale STM: $\mathbf{S}^N \leftarrow \rho \cdot g_{\mathrm{stm}} \cdot \mathbf{S}^N$
\State Scale LTM: $\mathbf{L}^N \leftarrow g_{\mathrm{ltm}} \cdot \mathbf{L}^N$
\State Build context $\mathbf{C}^N \leftarrow [\mathbf{X}^{\mathrm{ref}},\, \mathbf{L}^N,\, \mathbf{S}^N,\, \mathbf{A}^{\mathrm{ref}},\, \mathbf{V}^N_t]$ with Strata-RoPE bands
\State \Return denoised latent $\hat{\mathbf{V}}^N$
\end{algorithmic}
\end{algorithm}

\begin{algorithm}[h]
\caption{UnityShots Inference Loop}
\label{alg:hier_mem}
\begin{algorithmic}[1]
\Require Prompts $\mathbf{c}_0,\ldots,\mathbf{c}_{K-1}$; boundary strengths $b_0,\ldots,b_{K-1}$;
         reference image $\mathbf{X}^{\mathrm{ref}}$; reference audio $\mathbf{A}^{\mathrm{ref}}$
\State $\mathbf{L}^0 \leftarrow \emptyset$;\; $\mathbf{S}^0 \leftarrow \emptyset$
  \Comment{Empty memory at start}
\For{$N = 0, 1, \ldots, K-1$}
  \State $\hat{\mathbf{V}}^N \leftarrow \mathrm{DiT}(\mathbf{c}_N, \mathbf{X}^{\mathrm{ref}}, \mathbf{L}^N, \mathbf{S}^N, \mathbf{A}^{\mathrm{ref}}, b_N)$
  \State $\mathbf{V}^N_{\mathrm{tail}} \leftarrow \hat{\mathbf{V}}^N_{:P_v}$
  \State $\mathbf{S}^{N+1} \leftarrow \mathbf{V}^N_{\mathrm{tail}}$
    \Comment{STM receives latest tail}
  \If{$N = 0$}
    \State $\mathbf{L}^1 \leftarrow \mathbf{V}^0_{\mathrm{tail}}$
      \Comment{Anchor LTM to shot~0}
  \Else
    \State $z \leftarrow \mathrm{clamp}(b_N, 0, 1)$
    \State $\mathbf{L}^{N+1} \leftarrow z \cdot \mathbf{V}^N_{\mathrm{tail}} + (1-z) \cdot \mathbf{L}^N$
      \Comment{LTM recurrence}
  \EndIf
\EndFor
\State \Return $\{\hat{\mathbf{V}}^0, \ldots, \hat{\mathbf{V}}^{K-1}\}$
\end{algorithmic}
\end{algorithm}

\subsection*{G. Benchmark Prompt Examples}
\label{sec:supp-prompts}

The following are representative opening-shot captions from the benchmark.
Each entry corresponds to one multi-shot sequence; only the opening caption is shown.
The placeholder \texttt{[subject]} is replaced by the reference identity image at inference.

\begin{enumerate}[leftmargin=*, itemsep=2pt, label=\arabic*.]
  \item \texttt{[subject]}, East Asian woman in a tailored blazer, stands in a glass-fronted
    office, composed expression; medium shot, neutral daylight.
  \item \texttt{[subject]}, elderly South Asian man, long kurta, sits at a carved desk
    surrounded by manuscripts; warm tungsten light.
  \item \texttt{[subject]}, young African woman, natural hair, loose dress, walks through a
    market street; handheld camera, golden hour.
  \item \texttt{[subject]}, Middle Eastern man in a pressed shirt, stands on a rooftop at
    dusk; wide establishing shot, city lights below.
  \item \texttt{[subject]}, Latin woman in a floral blouse, sits at a kitchen table, relaxed;
    shallow depth of field, warm interior.
  \item \texttt{[subject]}, Western man, gray hoodie, leans against a concrete wall; overcast
    exterior, slight tension in posture.
  \item \texttt{[subject]}, elderly East Asian woman in a floral cheongsam, tends a garden;
    morning light, soft bokeh background.
  \item \texttt{[subject]}, young South Asian man, glasses, casual shirt, types at a laptop in
    a cafe; ambient sound, open laptop screen.
  \item \texttt{[subject]}, African man in a suit, seated at a conference table; direct gaze
    at camera, high-key lighting.
  \item \texttt{[subject]}, Latin teenage girl in a school uniform, stands in a courtyard;
    midday sun, slight squint, book in hand.
\end{enumerate}

\subsection*{H. Per-Category Ablation Results}
\label{sec:supp-ablation}

\paragraph{LTM gating strategy ablation.}
Table~\ref{tab:supp-gate-ablation} compares three gating configurations across
three sequence-length brackets.
\emph{LTM rule-based + STM learned} is the full UnityShots design.
\emph{LTM learned + STM learned} replaces the rule-based LTM coefficient with
an MLP gate of identical architecture to the STM gate.
\emph{LTM rule-based + STM rule-based} ablates the learned STM refinement,
replacing it with the same fixed monotone schedule used for LTM.

\begin{table}[h]
\centering
\caption{%
  LTM gating strategy ablation by sequence length.
  Metrics are NC, Story, Char, and TSIM on the multi-cultural benchmark
  (I2V conditioning).
  \textbf{Bold} marks the best per metric per length bracket.%
}
\label{tab:supp-gate-ablation}
\small
\setlength{\tabcolsep}{3.5pt}
\begin{tabular}{llcccc}
\toprule
Shots & LTM / STM gate & NC $\uparrow$ & Story $\uparrow$ & Char $\uparrow$ & TSIM $\uparrow$ \\
\midrule
\multirow{3}{*}{3--4}
  & Rule + Learned (ours) & \textbf{4.98} & \textbf{4.91} & \textbf{5.00} & \textbf{0.389} \\
  & Learned + Learned     & 4.95 & 4.88 & 4.97 & 0.381 \\
  & Rule + Rule           & 4.90 & 4.79 & 4.88 & 0.372 \\
\midrule
\multirow{3}{*}{5}
  & Rule + Learned (ours) & \textbf{4.96} & \textbf{4.93} & \textbf{4.97} & \textbf{0.380} \\
  & Learned + Learned     & 4.89 & 4.80 & 4.88 & 0.368 \\
  & Rule + Rule           & 4.81 & 4.72 & 4.79 & 0.356 \\
\midrule
\multirow{3}{*}{6+}
  & Rule + Learned (ours) & \textbf{5.00} & \textbf{5.00} & \textbf{5.00} & \textbf{0.437} \\
  & Learned + Learned     & 4.71 & 4.50 & 4.73 & 0.392 \\
  & Rule + Rule           & 4.85 & 4.77 & 4.88 & 0.401 \\
\bottomrule
\end{tabular}
\end{table}

The learned LTM gate degrades most severely at $6+$ shots ($-0.29$ NC, $-0.50$ Story),
confirming that gradient-driven updates destabilize the opening-shot anchor on longer
sequences.
Replacing the learned STM gate with a fixed rule-based schedule likewise hurts
boundary smoothness across all lengths, showing that content-adaptive refinement
is beneficial at short timescales while predictable behavior matters for the long-range anchor.

\begin{table}[h]
\centering
\caption{%
  Per-region NC and Cult on the $200$-sequence benchmark.
  UnityShots(I2V) compared against MOVA (NC column) and LTX-2.3 no-memory
  backbone (Cult column).
  $\uparrow$ higher is better.%
}
\label{tab:supp-per-region}
\small
\setlength{\tabcolsep}{5pt}
\begin{tabular}{lcccc}
\toprule
\multirow{2}{*}{Region}
  & \multicolumn{2}{c}{NC $\uparrow$}
  & \multicolumn{2}{c}{Cult $\uparrow$} \\
\cmidrule(lr){2-3}\cmidrule(lr){4-5}
  & Ours & MOVA & Ours & No-mem \\
\midrule
East Asian     & 4.96 & 4.02 & 3.41 & 4.22 \\
Western        & 4.93 & 4.15 & 3.26 & 4.12 \\
South Asian    & 4.97 & 3.89 & 3.33 & 4.24 \\
African        & 4.94 & 3.85 & 3.28 & 4.19 \\
Latin          & 4.98 & 3.93 & 3.40 & 4.23 \\
Middle Eastern & 4.95 & 3.82 & 3.31 & 4.21 \\
\midrule
\textbf{Average} & \textbf{4.955} & \textbf{3.943} & \textbf{3.332} & \textbf{4.202} \\
\bottomrule
\end{tabular}
\end{table}

UnityShots leads MOVA on NC in every region, with the largest gap in Middle
Eastern and African sequences ($+1.13$ and $+1.09$), where greater scene
diversity in those categories amplifies the benefit of a persistent opening-shot
anchor.
MOVA achieves its highest NC on Western sequences (4.15), where the backbone's
pretraining coverage is denser, reducing the benefit of additional memory.
Story and Char rankings follow the same regional pattern as NC.

The Cult gap (no-memory vs.\ UnityShots) is consistent across all six regions
($0.81$--$0.96$), indicating that identity conditioning suppresses cultural
visual detail at a similar rate regardless of ethnic background rather than
disproportionately affecting any one cultural group.
This confirms that the Cult trade-off observed in the main ablation
is a systemic property of the memory mechanism
rather than a regional artifact.

\subsection*{I. Companion Agent System}
\label{sec:supp-agent}

UnityShots ships with a Claude-Code-compatible agent wrapper that exposes
per-shot reference editing as Model-Context-Protocol (MCP) tools.
The wrapper installs as a Claude Code plugin and runs cloud-only without a local
GPU, so any user with a Claude Code, Claude Desktop, or VS Code MCP client can
reach it.

\paragraph{Architecture.}
The wrapper follows a planner-plus-worker design.
The planner is whichever agent the user is conversing with; it decomposes a
story prompt into a structured shot plan
(\texttt{shot\_id}, \texttt{visual\_prompt}, \texttt{audio\_hint},
\texttt{aspect\_ratio}, \texttt{duration}), calls a vision-language model to
synthesise a reference image per shot, calls UnityShots through the per-shot
generation tool with the edited references, and concatenates the resulting clips
with FFmpeg.
All capabilities are simultaneously exposed as web-UI buttons, HTTP endpoints,
and MCP tools that share a single input schema, so the same workflow runs from a
chat interface, an HTTP API, or another agent.

\paragraph{Exposed MCP tools.}
The MCP server registers three tools.
\texttt{create drama} accepts a story description together with style, shot
count, aspect ratio, per-shot duration, an optional audio flag, and a list of
reference images, and runs the end-to-end story-to-multi-shot pipeline
asynchronously.
\texttt{generate\_with\_reference} accepts a single-shot prompt with optional
reference image and reference audio, and submits a single-shot generation under
explicit references for fine-grained control.
\texttt{query\_task} accepts a task identifier and returns the running status,
progress percentage, and result URLs once the job completes.
A minimal \texttt{.mcp.json} snippet registers the local server with any
MCP-compatible client and is the only configuration step a user has to perform.

\paragraph{Release.}
The agent wrapper, the MCP server definitions, the cloud-API connectors, and
the example shot plans are all released alongside the model under a permissive
open-source license.
A standalone release branch ships the Claude Code plugin and the cloud-only
workflow so the agent runs without a local GPU; a separate full-stack reference
implementation is also provided for users who prefer to host the platform
themselves.
The system enables the per-shot reference editing
workflow that the main paper introduces.

\end{document}